\documentclass[journal]{IEEEtran}
\usepackage{cite}

   \usepackage[pdftex]{graphicx}

\usepackage{caption}
\usepackage{subcaption}
\usepackage{moreverb,url}
\usepackage[colorlinks,bookmarksopen,bookmarksnumbered,citecolor=red,urlcolor=red]{hyperref}

\usepackage{amsmath}

\begin{document}

\title{Mixed-Initiative variable autonomy for remotely operated mobile robots}
%
%
%

\author{Manolis~Chiou,
        Nick~Hawes,
        and~Rustam~Stolkin
\thanks{M. Chiou and R. Stolkin are with the Extreme Robotics Lab, University of Birmingham, UK}
\thanks{N. Hawes is with the Oxford Robotics Institute, Department of Engineering Science, University of Oxford, UK}
}
\maketitle

\begin{abstract}
This paper presents an Expert-guided Mixed-Initiative Control Switcher (EMICS) for remotely operated mobile robots. The EMICS enables switching between different levels of autonomy during task execution initiated by either the human operator and/or the EMICS. The EMICS is evaluated in two disaster response inspired experiments, one with a simulated robot and test arena, and one with a real robot in a realistic environment.

Analyses from the two experiments provide evidence that: a) Human-Initiative (HI) systems outperform systems with single modes of operation, such as pure teleoperation, in navigation tasks; b) in the context of the simulated robot experiment, Mixed-Initiative (MI) systems provide improved performance in navigation tasks, improved operator performance in cognitive demanding secondary tasks, and improved operator workload compared to HI. Results also reinforce previous human-robot interaction evidence regarding the importance of the operator's personality traits and their trust in the autonomous system. Lastly, our experiment on a physical robot provides empirical evidence that identify two major challenges for MI control: a) the design of \textit{context-aware} MI control systems; and b) the \textit{conflict for control} between the robot's MI control system and the operator. Insights regarding these challenges are discussed and ways to tackle them are proposed.  
\end{abstract}


%
\IEEEpeerreviewmaketitle

\section{Introduction}
In the aftermath of the 9/11 World Trade Center (WTC) terrorist attack, several robots were used in an Urban Search and Rescue (USAR) operation. The robots' tasks were to inspect areas beneath debris and rubble, or enter confined spaces (e.g. dangerous voids), that humans or dog rescuers could not enter. Robot operators were working under stressful and performance-degrading conditions such as cognitive fatigue, sleep deprivation, and minimum sensor information (e.g. caused by dust or poor lighting) which led to mistakes in robot operation \cite{CasperMurphy911}.

On March 11th 2011 a tsunami hit Fukushima Daiichi nuclear power plant causing three nuclear meltdowns and the release of radioactive material. In an effort to initially assess and contain the situation, military-grade robots were used to inspect the buildings and the reactors for any damage, radiation, and other potentially hazardous conditions. These robots had to be operated by the nuclear power plant workers, after only brief training sessions. A typical scenario involved those newly trained operators remotely controlling the robots for many hours while wearing a hazmat suit, making operation a very challenging task. Additionally, these operators had to face a number of psychologically challenging and stressful situations (e.g. dosimeter alarms going off), as famously reported by one of the robot operators in his blog\footnote{IEEE Spectrum, Accessed: 2020-06-26: \url{http://spectrum.ieee.org/automaton/robotics/industrial-robots/fukushima-robot-operator-diaries}}.

Several years after these disasters, many of the initial challenges and shortcomings of robot use in disaster response remain. Despite the need to make the use of robots in safety and time critical applications (e.g. USAR, nuclear decommissioning, and hazardous environment inspection) easier, robots used are still predominately teleoperated, with little or no autonomy used to assist the human operator. There are three main reasons why autonomous systems are not used in these settings. First, these tasks often involve environments that are highly unstructured and changing, e.g. a partially collapsed building. Second, the nature of these tasks requires specific human abilities such as critical decision making based on incomplete information (e.g. determining if a victim dead or alive; and risk assessment of certain actions); or communication with victims in Search and Rescue (SAR) \cite{Dole2015}. Third, high consequence industries tend to be conservative and therefore do not trust autonomous systems. For example in the Fukushima Daiichi nuclear accident, such lack of trust led to the robots being deployed via pure teleoperation, despite having various autonomous capabilities \cite{Nagatani2013}. For these reasons, current deployment of such robots always requires a human in the loop \cite{Murphy2004}.


Equipping robots with autonomous capabilities can potentially tackle some of the intrinsic difficulties of teleoperation. Several field studies have pointed out the need for robots that actively assist operators (e.g. in the 9/11 World trade center \cite{CasperMurphy911}; the DARPA robotics challenge \cite{Yanco2015DARPA, Norton2017}; and other studies e.g. \cite{Murphy2005a, Music2017}). What is required is a human-robot team that benefits from the capabilities of both agents, whilst counteracting the weaknesses of each. Such systems offer the potential to assist a human operator who may be struggling to cope with issues such as high workload, intermittent communications, operator multitasking, fatigue, and sleep deprivation. For example, a human operator might need to concentrate on a secondary task while temporarily delegating control to the robot to navigate autonomously. This is something very common as robot operators have to convey situational awareness information, e.g. to SAR task force team mates \cite{Murphy2005a,Burke2004a}.

Our research addresses the use of \emph{variable autonomy} as an approach to blending the capabilities of humans and robots. A variable autonomy system is one in which control can be traded between a human operator and a robot by \emph{switching between different Levels of Autonomy} \cite{Chiou2016}. Levels of Autonomy (LOAs) refer to the degree to which the robot, or any artificial agent, takes its own decisions and acts autonomously  \cite{Sheridan1978}. In the case of robotics, LOAs can vary from the level of pure teleoperation (human has complete control of the robot), to the other extreme which is full autonomy (robot has control of every capability), within a single robot.

This article addresses the problem of dynamically changing LOA on-the-fly (i.e. during task execution) using either \emph{Human-Initiative} (HI) or \emph{Mixed-Initiative} (MI) control. HI refers to a system where the human operator is solely responsible for switching LOA based on their judgement. MI refers to a system where both the human operator and the robot's control switcher have authority to initiate actions and to change the LOA. How best to dynamically switch LOAs in order to improve system performance is a challenging and open problem that remains largely unexplored in the literature.

The hypothesis explored in this article is that timely switching between different LOAs during task execution (e.g. during navigation) can improve task performance and will enable the system to overcome various performance-degrading factors. This is compared to robotic systems in which LOAs cannot switch on-the-fly. More specifically we investigate the capacity of the human operator (based on their judgement) and the robot's MI control switcher (based on an online performance metric) to use HI and MI control in order to improve performance compared to pure teleoperation or autonomy. In the rest of this article, when we use the term ``robot'', we use it for the physical robot hardware. The term Expert-guided Mixed-Initiative Control Switcher (EMICS) will refer to the robot's MI control switcher, i.e. the autonomous software system that initiates the LOA switching. The terms autonomy and teleoperation will refer to the individual LOA in terms of navigation.

\subsection{Scope of current research}

Our initial work \cite{Chiou2015} demonstrated that conducting variable autonomy experiments is challenging due to the high number of intrinsic confounding factors that can lead to large variances in the results. Such confounding factors include individual differences in personality traits; experience in operating robots or playing video games; and map exploration strategies. In subsequent work we contributed a systematic variable autonomy experimental framework to avoid these issues~\cite{Chiou2016}. In \cite{Chiou2016}, human operators used an Operator Control Unit (OCU) and HI control to navigate a remotely controlled mobile robot in a maze-like test arena. The OCU was a laptop, joypad, mouse, and a screen showing the control interface (i.e. map, robot status, and video feedback from robot's camera). The HI control switcher allowed operators to switch between two different LOAs on the fly: an autonomy LOA, in which the operator could give a navigational goal by clicking on the interface, causing the robot to navigate autonomously towards that point; and a teleoperation LOA, in which the human operator manually controlled the robot by using the joypad. The results showed that that HI variable autonomy can overcome various performance-degrading factors and outperform both pure teleoperation and pure autonomy in various circumstances. Additionally, in \cite{Chiou2016_AAAI} we explored the HRI aspects of the human operator interacting with and exploiting the HI control switcher. It provided evidence that operators' interactions with the variable autonomy control switcher (e.g. frequency of LOA switching; preferred LOA) are not necessarily influenced only by task performance. 

In this paper we advance variable autonomy research by moving from HI to MI robotic systems. We also provide a complete picture of research regarding robots which dynamically switch LOA and make the case that HI or MI control provides advantages over single modes of operation (e.g. pure teleoperation or pure autonomy). This is done through the design and evaluation of an MI control switcher (i.e. the EMICS).

\subsection{Contributions}
This article makes the following contributions. It presents:

\begin{itemize}

	\item statistically-validated empirical evidence that HI systems outperform single mode of operation systems (e.g. pure teleoperation) in navigation tasks, based on real robot USAR-inspired experiments.
    
    \item a framework and guidance for designing robotic MI control systems. More specifically, this paper proposes the design of \emph{expert-guided MI control switchers}.
    
	\item an MI control system in which both the operator (based on judgement) and the robot's EMICS (based on an online performance metric) are able to switch between different LOAs. Our approach in designing the EMICS makes use of: expert knowledge; an online performance metric; and simplified contextual knowledge to infer if a LOA switch is needed.

\end{itemize}	

We also provide, a rigorous evaluation plus statistically-validated empirical evidence regarding the potential advantages of the MI control in various circumstances, as compared to HI and teleoperation. Our simulated robot experiments demonstrate the following advantages: a) improved performance in navigation tasks; b) improved operator performance in cognitively demanding secondary tasks such as the mental rotation of 3D objects; c) a reduction in operator workload. In addition, this article contributes an analysis of the interaction of a human operator with an MI system and reports on metrics such as time spent in each LOA; frequency of LOA switches; perceived workload; and their correlation with system performance. Finally, based on a physical robot experiment, we identify and provide empirical evidence and insights into two major challenges for MI control: the design of context-aware MI control systems; and the conflict for control between the operator and the MI control system.

\section{Background and related work} 

In this section, we discuss research on strategies for switching LOAs. More specifically, we identify gaps in the literature regarding three key aspects: a) conducting rigorous experiments on LOA switching; b) LOA switching initiated by the human operator (namely HI); and c) LOA switching initiated by the robot's MI control switcher or the operator (namely MI). The focus of this paper is MI control and thus the literature relevant to MI robotic systems will be reviewed in detail. Literature on HI control is briefly discussed for completeness. For further information on experimental frameworks and HI please refer to our previous work \cite{Chiou2015,Chiou2016,Chiou2016_AAAI}.

\subsection{Human-Initiative variable autonomy}

A Human-Initiative (HI) variable autonomy system is a system in which the human operator can dynamically switch between different LOAs (e.g. between teleoperation and autonomy) \cite{Chiou2016}. In such systems only the human operator has the authority to initiate LOA switches based on their judgement. Many of the systems found in literature, such as \cite{Goodrich2001}, are restricted to an initial LOA choice as they cannot change LOA on the fly. Other systems (e.g. \cite{Baker2004b, Ibanez-Guzman2004, Finzi2005}) aid the operator's judgement by suggesting potential changes in the LOA. Other SAR-inspired studies (e.g. \cite{Marble2004}) are mostly focused on evaluating the usability of such HI systems. 

\subsection{Human-Robot Interaction with LOA switching robots}

Human-Robot Interaction (HRI) in a LOA switching system remains mainly unexplored in the prior literature. Previous variable autonomy studies investigated the human operator's interaction with a robotic system, but were restricted to exploring a single LOA \cite{Bruemmer2005, Carlson2008}. Other studies, \cite{Baker2004b,Ibanez-Guzman2004,Marble2004} did not present any data on the operator's interaction with the LOA switching controller. In \cite{Chiou2016_AAAI}, an analysis of the ways in which human operators interact with, and exploit the capabilities of, a HI robotic system with dynamic LOA capabilities is reported. In this article we report on the HRI between the operator and the EMICS.

\subsection{Definition and taxonomy of Mixed-Initiative control}
\label{section:MI-taxonomy}

In \cite{Jiang2015} Jiang and Arkin define MI control in the context of human-robot teams as:

\begin{quotation}
	``A collaboration strategy for human-robot teams where humans and robots opportunistically seize (relinquish) initiative from (to) each other as a mission is being executed, where initiative is an element of the mission that can range from low-level motion control of the robot to high-level specification of mission goals, and the initiative is mixed only when each member is authorized to intervene and seize control of it.''
\end{quotation}

They also present the first taxonomy for MI robotic systems. Their taxonomy has three dimensions.

The first dimension is \textbf{span-of-mixed-initiative} which characterizes the control elements (initiatives) in which both agents are capable of initiating actions. The system we present in Section \ref{fuzzy_controller} is \textit{mostly-joint} (a term from this dimension in ~\cite{Jiang2015}) since both agents have initiative over two of the control elements: navigation execution and LOA switch. The second taxonomy dimension is \textbf{initiative reasoning capacity} which characterizes the ability of an agent to reason about taking the initiative. Our system is \textit{deliberative} since it has the ability to reason about initiating actions deliberately based on an online performance metric and simplified context awareness. The final dimension of the taxonomy is \textbf{initiative hand-off coordination} which characterizes the strategies used by the system when shifting initiative from one agent to the other. Our system is characterized as \textit{explicitly-coordinated} in this dimension.

\subsection{Mixed-Initiative control systems}
\label{lit_review_MI-systems}

Given the comprehensive taxonomy in \cite{Jiang2015}, in this survey we review systems that have some form of MI control (i.e. both human and operator can initiate actions or LOA switches). Shared control is a term often used generally to describe systems in which the human cooperates at some level with the robot's Artificial Intelligence (AI) controller; or is used specifically to describe research and systems in which some form of input mixing or blending between the robot's AI controller and/or the operator's commands is used. In this paper, to avoid confusion, we will use this term to explicitly refer to research involving some form of input mixing. Shared control from our perspective is a LOA that can fall under the banner of MI control. Any mixed initiative behaviour is restricted within a shared control LOA. This means that the robot's AI controller can only take the initiative to blend its navigation control input with the one from the operator, in order to improve the control output. Similarly, in safeguard teleoperation (another form of shared control), the AI controller reactively takes the initiative to prevent collisions. In both cases no LOA switching takes place. Also in most cases the human operator does not have any initiative in choosing how to blend the control inputs. In Section \ref{section:shared_control_relevance} further discussion can be found on how our system relates to shared control and input blending.

Nielsen et al. \cite{Nielsen2008} conduct experiments using multiple LOAs. The LOA is chosen during the initialization of the system and cannot change on the fly. Moreover, similar to shared control, the AI controller has reactive initiative inside a specific LOA to prevent collisions. Multiple LOAs (teleoperation, safe mode, shared mode, autonomy) are tested in \cite{Bruemmer2005}. In shared mode the robot's AI controller drives autonomously while accepting interventions from the operator. In safe mode the AI controller takes initiative to prevent collisions. Few et al. \cite{Few2006} present a control mode in which the operator gives directional commands using a joypad to adjust the robot's navigation. The system offers initiative which depends on the frequency of the operators interaction. Rigter et al \cite{Rigter2020} propose a contextual multi-armed bandit framework for learning policies that enable a control switcher to choose between LOAs with the goal to minimize human interventions e.g. recovering the robot from failures.

Gombolay et al. \cite{Gombolay2015,Gombolay2017} use an industrial assembly scenario to investigate how varying degrees of task allocation authority in physical collaboration between the humans and the robot affect the human-robot team's task performance, SA and task work-flow. Their system has three different LOAs that can be chosen in the beginning of the scenario: a) the human decides how to allocate the tasks; b) the human decides which tasks they will perform, while the AI controller allocates the remaining tasks to itself and to other humans; c) the AI controller has full authority on task allocation.

Bruemmer et al. \cite{Bruemmer2003b} present a theoretical, multiple LOA, MI system. This system is based on theories of robot behavior (human understanding of the robot) and human behavior (robot understanding of the human). The latter proposes gathering readily available non-intrusive workload cues from the operator as an indication of poor performance. More specifically, it proposes the use of the frequency of human input and the number and kind of dangerous commands issued by the operator as performance indicators. This can provide the robot's MI control system with the capacity to initiate switches between the different LOAs. Adams et al. \cite{Adams2004} propose a MI robot control architecture which relies on the detection of the operator's emotional state. Initiative is mixed in all the levels of the system, i.e. in setting goals and constraints, planning and execution. Changes in control are initiated based on the operator's sensed state (e.g. boredom, stress, drowsiness, engagement). 

For multi-robot systems, variable autonomy often lies at a higher level of abstraction compared to single robot systems. Manikonda et al. \cite{Manikonda2007} describe a multi-robot MI control system and testbed for human-robot teams in tactical operations. The agents in the system share information via a common world model. Based on this information they are able to initiate modifications to their goals and associated roles in the team. In \cite{Hardin2009} a MI approach is proposed in a multi-robot search task. Robots are equipped with the ability to initiate changes in their respective search areas (e.g. size of search area). These changes are reactively triggered by specific events, e.g. the human operator has identified an item of interest. In \cite{Liu2016} and \cite{Hong2018} a hierarchical reinforcement learning architecture is proposed for semi-autonomous control of multiple robots by one operator. It is focused on task allocation between robots. Additionally, when a task is failing, the system asks for operator's help.  

In summary, MI robotic systems found in the literature often offer limited initiative inside a predefined LOA. Some of the MI systems proposed are either purely theoretical, or not experimentally evaluated. Moreover, initiative actions from the robot's AI controller are often based on reacting to sensor input (e.g. obstacles) rather than online task performance metrics. In the case of multi-robot systems, the MI lies in a higher level of abstraction, making assumptions about other layers, e.g. navigation. 

The work presented in this article aims to address some of the gaps in the literature by focusing on the problem of switching LOA during task execution. This is achieved by designing a MI control switcher which relies on an online task performance metric. Additionally, it aims to show benefits to human operator cognitive workload and benefits to tasks performance (e.g. navigation, spacial awareness tasks etc) in a robotic system that initiates dynamic LOA switches. 

Lastly, some of the literature lack the use of a rigorous experimental framework, e.g. absence of standardized training for the human operator; absence of statistical analysis; protocols not explained. In this paper we aim to address this shortcoming by using a rigorous experimental framework which consists of: a) appropriate statistical analysis; b) clarity on assumptions and hypotheses; c) precise and detailed descriptions of the experimental protocols followed; and d) a formalized and repeatable experimental paradigm. Because of the nature of this research (i.e. human-robot systems in complex problems and environments) it is very difficult for this or other related studies to fully control for every possible confounding factor. Here, we aim for a balance between realism (i.e. realistic tasks and scenarios) and meaningful scientific inference (i.e. controlled experiment).


\section{An expert-guided framework for designing Mixed-Initiative control systems}

The fundamental problem in MI control is how to allow LOA switching by either the operator or the robot's MI control switcher in order to improve the human-robot system's performance. Without loss of generality, in the rest of this paper we assume a human-robot system which has two LOAs: one in which a human has most of the control (i.e. some form of teleoperation), and one in which an AI controller has most of the control (i.e. some form of autonomous control). However the descriptions in the following can be extended to the case where other LOAs exist, e.g. alternate autonomous or shared control LOAs.

In MI control the robot, meaning the hardware, can be seen as a resource with two different agents having control rights: the human operator and the MI control system. At any given moment, the most capable agent should take control. Hence, of particular importance is the ability for each agent to diagnose the need for an LOA change, and to take or relinquish control successfully.

An operator's LOA switches are based on judgement. As demonstrated by the HI results in \cite{Chiou2016}, given sufficient understanding of the system and the situation humans are able to determine when they need to change LOA. More specifically, analysis of the HI results from our previous work \cite{Chiou2016_AAAI} revealed that operators switch LOA based on three factors: a) preferred LOA; b) context; and c) performance degradation. The preferred LOA is the LOA the participants tend to instinctively return to. Their preference may be based on a number of task- or system-specific factors (e.g. trust in the control software) and personal traits (e.g. preference to be in control). In context-sensitive LOA switches, operators are able to evaluate the current context and infer if a LOA switch is needed. For example they can change control preemptively as they predict performance degradation in a given situation. An example is a situation in which noise starts to appear in the sensors, thus they change to teleoperation as they predict degradation in the autonomous controller's performance.

In order to enable the MI control system to automatically take control when it is needed, and to relinquish control when under-performing, it is necessary to provide it with a mechanism to detect when the performance of the human-robot system is degraded. For example, the MI control switcher could initiate a switch to a LOA that offers increased autonomous capabilities in situations where the human operator is too preoccupied with the primary cause of their performance degradation to voluntarily switch control to the robot's MI control system (e.g. the autonomous navigation controller).

To this end we propose an \emph{expert-guided approach} to the design of MI control switchers. For a given task, we assume the existence of a \emph{task expert} that can provide the expected task performance for the human-robot system in the absence of performance-degrading or other unexpected factors. This expert is needed in order to provide a reference point for behavior and performance, which might be different from the actual runtime performance and behavior of the system. Based on this assumption, the core idea is that the EMICS should compare the current performance of the system with the expert performance. This comparison yields an online \textit{task effectiveness} metric, expressing how effective the system is performing its given task relative to the performance of the expert. If this metric indicates that the current task performance is worse than the expert expects, then the EMICS should initiate a LOA switch. Such experts can be created in many ways, e.g. through simulation, assuming no noise, learning by demonstration, etc. Depending on the application and the implementation, such experts can represent an optimal, close-to-optimal, heuristic, or optimistic performance reference point for the actual system performance to compare against at runtime.

The application of our expert-guided MI approach to a given human-robot team and task relies on the following assumptions: a) the human operator is willing to be handed control and to hand over control based on the initiative of the EMICS; b) the agent to which the control will be handed (i.e. either human or the MI control system) is capable of correcting the task effectiveness degradation \cite{Chiou2015}; and c) the system is equipped with an expert controller providing information on how (given the assumptions made in the design of the expert) the system should be performing in a given task. 


\subsection{An expert-guided Mixed-Initiative control switcher for navigation}

Many of the applications for remotely operated robots involve navigation. Therefore for the remainder of this article we focus on the problem of MI control for navigation, specifically the task of the moving the robot to goal location as quickly and as safely as possible.

Creating an EMICS for navigation requires an online task effectiveness measure for navigation. Hence, we have chosen to use the \textit{goal-directed motion error} (referred to as ``error'' for the rest of the article). This is the difference between the robot's current motion and the motion of the robot required to achieve its goal (reach a target location).

Such a motion error can be trivially provided by taking the difference between the robot's current direction of travel and the direction directly towards location of the navigation target, however this does not account for environment structure. To provide more environment context to the EMICS we extract an expert performance measure from a concurrently active navigation system which is given an idealised (unmapped obstacle- and noise-free) view of the robot's world. In a real scenario (e.g. the robot has to inspect a building for damage) such view or map can be provided to the system as prior knowledge, e.g. taken from the building's plan. This expert uses the navigation planner from the Robot Operating System (ROS) navigation stack \cite{Marder-Eppstein2010}. More specifically, the global shortest feasible path is calculated based on Dijkstra's algorithm \cite{Dijkstra1959}, while the local path and optimal velocities are calculated using the dynamic window approach \cite{dwa_planner}. The input to the navigation planner is the robot's location (from its localisation system), its target location, plus only the static map of the environment. Using this, the expert continually reports the velocity with which the robot should be moving towards its goal. By omitting dynamic obstacle maps from the planning process (i.e. additional obstacles detected by sensing elements of the environment not present in the map) the velocity suggested by the expert can be seen as an upper bound on the performance of the system. In essence, the expert provides an optimistic model of possible robot behavior.

\begin{figure}
	\centering
	\includegraphics[width=0.5\columnwidth]{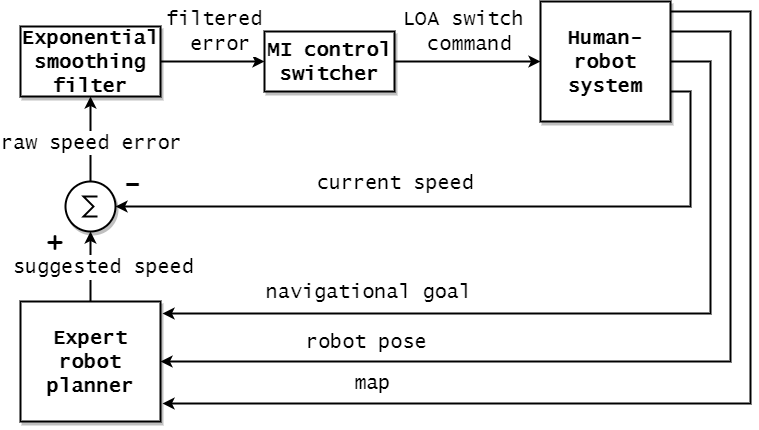}
	\caption{The block diagram of the expert-guided MI control system. Given a navigation goal; the current pose of the robot; and the map (or a known surrounding area); the expert planner yields an ideal suggested speed. This speed denotes how fast the robot should be moving towards achieving the navigation goal. Then this speed is compared with the current speed of the robot towards that goal to calculate the raw performance error. The MI control switcher (i.e. the EMICS) decides on switching LOA based on the filtered error.} 
	\label{fig:block_diagram}
\end{figure}

From the expert-reported velocity we extract the speed in the $x$ axis (i.e. the axis denoting the forward/backward motion of the robot). This denotes the speed which the robot should be moving along the optimal path towards the goal. To create an error measure the expert's $x$ speed is constantly compared with the current $x$ speed of the robot. The difference these values is the \textit{goal directed motion error} that we use in the control switchers presented in the remainder of this article. Please refer to Figure \ref{fig:block_diagram} for the block diagram of the system.

In addition to the general assumptions described in the previous section, our navigation-based expert MI approach assumes that: a) the region surrounding the robot is mapped; b) the robot's control system has access to a target location for navigation; and c) the system is equipped with a navigation expert planner capable of reliably computing both an idealised path and velocities towards a navigation goal. The assumptions a and b represent the knowledge of the task, its environment, and what it is to be achieved. This knowledge is required by the expert planner in order to infer the idealised path and velocities. In a real scenario the robot operator can have such prior knowledge which can use as input to the robot. For example, the operator can provide a map to the robot with the floor plan (e.g. of the building to be explored, taken from the authorities or mapped by a UAV) and the coordinates of a point of interest (e.g. the location of a pipe suspected to be leaking or the potential location of victims).

\subsection{Threshold Mixed-Initiative control switcher}
\label{threshold_controller}

Using the above approach we created two different MI control switchers. Our first approach was to create a threshold control switcher using the \textit{goal directed motion error} metric. This control switcher observes the system's performance over a period of time. If performance drops below a given threshold then a switch in the LOA is initiated regardless of which agent is in control. To implement this approach, the goal-directed motion error must satisfy the following requirements: a) frequent, brief changes in error should be treated as noise; and b) the final error signal to be used should express the accumulated error over time. The second requirement comes from our model since it assumes that the performance drop is observed over a short period of time before any initiative takes place.

The exponential moving average (EMA) algorithm (commonly known as Brown's simple exponential smoothing \cite{Brown1963}) applied (see equation \ref{eq:avg}) to the raw error (equation \ref{eq:error}), satisfies both of these requirements. The raw error equation follows:

\begin{equation}\label{eq:error}
e_{t} = s_{E} - s_{R}
\end{equation}

The raw error $e_{t}$ is the difference between the ideal speed from the expert planner $s_{E}$ and the current linear speed $s_{R}$ of the robot moving towards the goal. The equation for the smoothed error using the EMA follows:

\begin{equation}\label{eq:avg}
E_{t} = \alpha \cdot e_{t} + (1-\alpha) \cdot E_{t-1}
\end{equation}

The term $E_{t}$ refers to the final accumulated and smoothed error. The term $e_{t}$ refers to the current error observation. The smoothing factor $\alpha$ controls the weight of the most recent (i.e. current) observation and thus the time window in which the error will be accumulated. 

The threshold control switcher uses the filtered error $E_{t}$ to initiate LOA switches. When this error exceeds a given threshold, the control switcher infers that the current agent in control is under-performing. Thus, it initiates a change in the LOA. To calculate the error threshold and the smoothing factor $\alpha$ we used the following procedure. We replayed the robot motion for all operator trials from the HI experiment in \cite{Chiou2016} through the control switcher with a given threshold and $\alpha$ both selected via grid search. For each trial this yielded a set of LOA changes proposed by the control switcher. We compared this set of LOA switches to the LOA switches initiated by the operators in the previous HI data. We then used the  cost function in Equation \ref{eq:cost_individual} to calculate the cost of a particular parametrisation of the control switcher for each individual trial. The final parameters chosen were the parameters that minimized the sum of costs for all participants. The cost function for each individual trial was:

\begin{equation}\label{eq:cost_individual}
j = \sum{|\hat{t^{i}_k} - t_k|} + c \cdot p
\end{equation}

Put simply, equation \ref{eq:cost_individual} expresses the LOA switch prediction error from the control switcher on HI data, with the addition of a penalty term for every prediction that does not match (i.e. false positives) where $\hat{t}$ is the time-stamp of the predicted LOA switch and $t$ is the time-stamp of the operator's actual switch. The subscript $k$ denotes a specific time-stamp, where $c$ is the number of predictions that do not match operator's switches, and $p$ is a cost penalty. A non-matching prediction is defined as a prediction that falls outside of a small time window around any of the operator's LOA switches. The assumption is that these predictions do not correspond to any actual operator's LOA switches and thus they get penalized. 

The parameters calculated with the grid search were $\alpha = 0.06$ and error $threshold = 0.07 m/sec$. The $\alpha$ calculated here represents a trade-off between fast reactions to large current errors and filtering noise (i.e. sudden instantaneous errors). These two extremes (i.e. fast reactions versus react cautiously to accumulated error) are both included in realistic navigation tasks similar to the ones in our experiments, as they can both occur during task execution. Hence, the $\alpha$ value does not need to be re-derived for every scenario as long as the scenario is of the same nature and not a specialized case of navigation. For this reason, the sensitivity of the control switcher to react and switch LOA should be primarily governed by how the error value is managed.

Using these parameters with the threshold control switcher we conducted a pilot experiment with a small number of participants. It rapidly became clear that this control switcher did not work well. Poor performance was characterized by excessive LOA switching (i.e. the control switcher was oversensitive in detecting performance drops). As a result the control switcher was intrusive and impractical. This led us to design a new expert knowledge control switcher based on fuzzy logic, as described in the next section.

\subsection{Fuzzy Mixed-Initiative control switcher}
\label{fuzzy_controller}

 A bang-bang (also known as on-off controller) Mamdani type fuzzy controller \cite{Mamdani1975,Nagi2009} was designed to address the limitations of the threshold control switcher. The fuzzy lite \cite{fuzzylite} C++ library was used for the task. The repository\footnote{Extreme Robotics Lab GitHub repository: 
\url{https://github.com/uob-erl/fuzzy_mi_controller}} containing the ROS code for the MI control switcher and any code necessary to replicate the experiments described in this paper, is provided under MIT license.
 
In this case, the bang-bang control switcher has two states: a) switch LOA; b) do not switch LOA. This control switcher goes beyond the threshold control switcher by utilizing expert knowledge in three ways: a) by defining fuzzy sets for the goal directed motion error, instead of a fixed threshold; b) by incorporating very simple context information; and c) by taking informed decisions on when to initiate a LOA switch based on a fuzzy logic rule base. The hypothesis is that the fuzzy logic EMICS will yield smoother transitions between LOA switching decisions. Thus, will not be as intrusive (i.e. reactive) in switching LOA compared to the simple threshold control switcher.

In fuzzy logic a linguistic variable associates a linguistic concept and the fuzzy set representing that concept to a numerical value.
Our fuzzy control switcher has two variables as inputs. The first linguistic input variable is ``error'' that denotes the filtered error. This is the filtered error from the exponential moving average (EMA) as in the previous threshold control switcher. The smoothing factor $\alpha$ used in the EMA filter, was the one found by the parameters search for the threshold control switcher. The second input variable is ``speed'' that denotes the speed at which the robot is currently moving. 

\begin{figure*}
		\centering
		\begin{subfigure}[b]{0.47\textwidth}
			\centering
			\includegraphics[width=\textwidth]{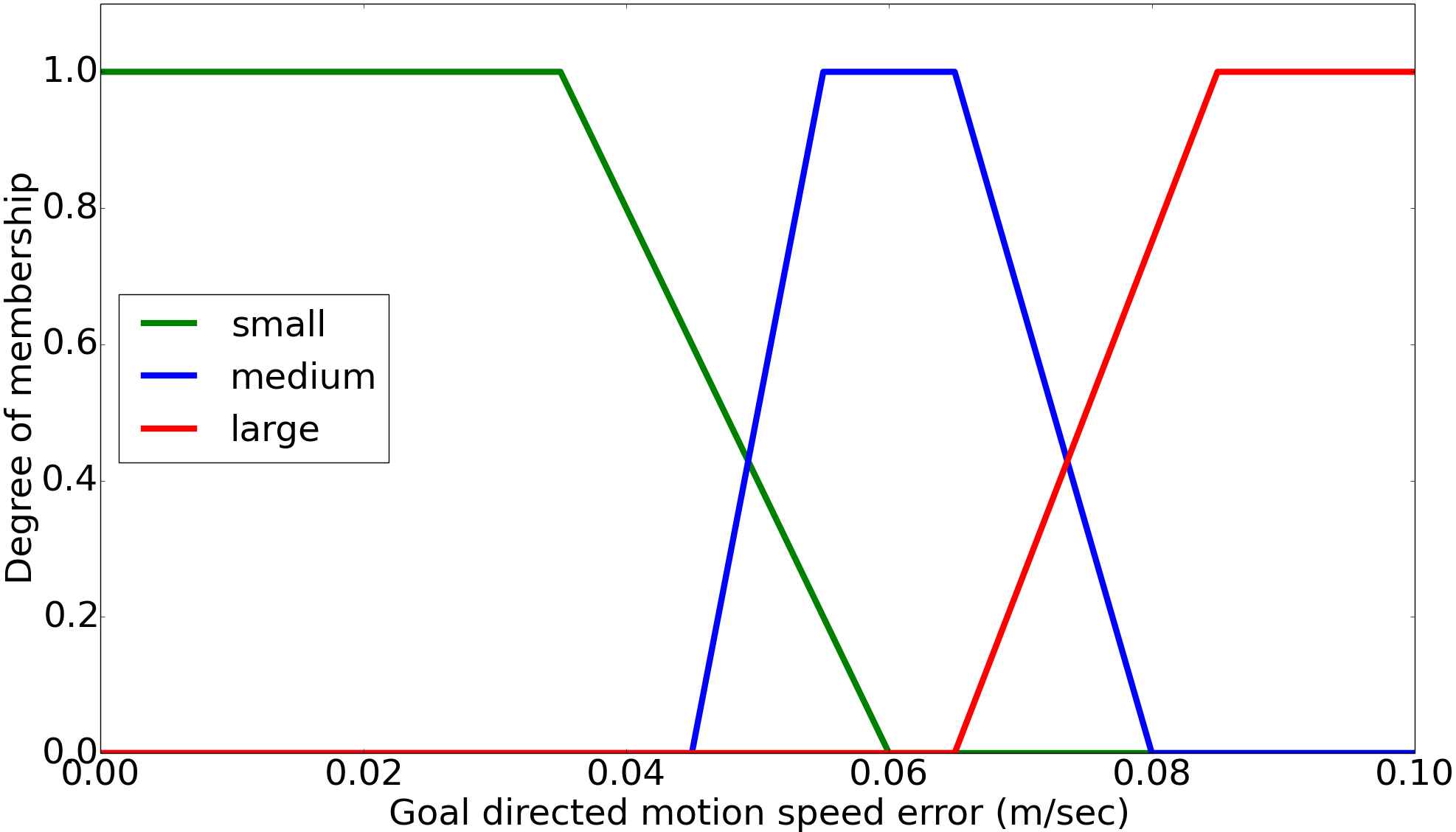}
			\caption{}
			\label{subfig:fuzzy_error}
		\end{subfigure}
		\hfill
		\begin{subfigure}[b]{0.47\textwidth}
			\centering
			\includegraphics[width=\textwidth]{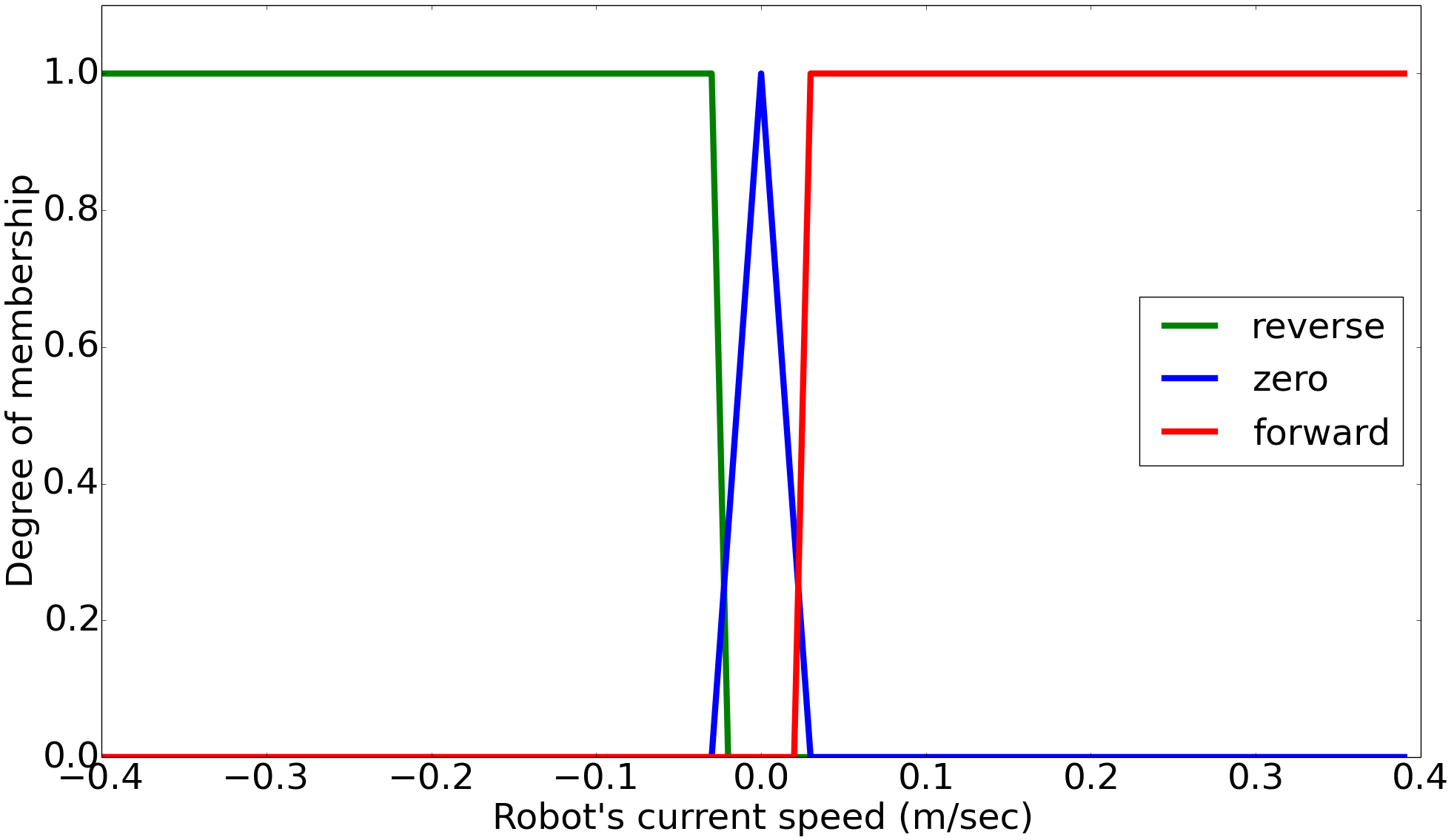}
			\caption{}
			\label{subfig:fuzzy_speed}
		\end{subfigure}
		\hfill
		\caption{\textbf{\ref{subfig:fuzzy_error}:} Membership functions for the linguistic input variable ``error". \textbf{\ref{subfig:fuzzy_speed}:} Membership functions for linguistic input variable ``speed".}
		\label{fig:fuzzy_input}
\end{figure*}

For the linguistic input variable $x_1 = ``error"$ the universe of discourse (i.e. the range of all possible values for an input) is $X_1 =[0,0.1] m/sec$. A value of $0 m/sec$ means that no goal directed motion error exists, i.e. the robot is making progress towards the goal without any performance degradation. The value of $0.1 m/sec$ is the maximum error, meaning that the robot is not progressing towards the goal (e.g. robot is idle). This is despite the maximum speed of the robot being $0.4 m/sec$. Due to physical and mechanical constrains (e.g. acceleration limits) the maximum new speed that can be given to the robot at any time, cannot have a difference with the current speed more than $0.1 m/sec$. For the linguistic input variable $x_2 = ``speed"$ the universe of discourse is $X_2 = [-0.4,0.4] m/sec$. The value of $0.4 m/sec$ is the maximum speed of the robot. The value of $-0.4 m/sec$ is the maximum reverse speed of the robot. 

The input variable ``error'' (i.e. $x_1$) is mapped into three linguistic values (i.e. three fuzzy set membership functions, see Figure \ref{subfig:fuzzy_error}). The set $A_1^i$ for input $x_1$ can be defined for the following linguistic values: $A_1^i = [A_1^1 = small, A_1^2 = medium, A_1^3 = large]$. The error threshold calculated in our previous threshold control switcher was used to denote the fuzzy linguistic value ``large". In essence what operators consider to be a large error to justify a LOA switch, is encoded into the fuzzy control switcher. This knowledge was extracted by using a grid search algorithm on HI data as described in Section \ref{threshold_controller}. The values and membership functions for ``small'' and ``medium'' were heuristically chosen in order to smoothly overlap throughout the universe of discourse (see Table \ref{table:membership_functions}). This is a common practice when designing fuzzy controllers. 

The input variable ``speed'' (i.e. $x_2$) is mapped into three linguistic values (see Figure \ref{subfig:fuzzy_speed}). The set $A_2^i$ for input $x_2$ can be defined for the following linguistic values: $A_2^i = [A_2^1 = reverse, A_2^2 = zero, A_2^3 = forward]$. The value ``reverse'' denotes that the robot's speed is negative, which means the robot is reversing. The value ``zero'' denotes that the robot is idle and not moving. The value ``forward'' denotes that the robot is moving forward (see Table \ref{table:membership_functions}).

The fuzzification process transforms the crisp values of the inputs into fuzzy values. This is achieved using the fuzzy membership functions described above and in Table \ref{table:membership_functions}. Then, a set of fuzzy rules is applied to the fuzzy inputs (see Table \ref{table:rule_base}). The following standard (i.e. commonly used) operators are used in the fuzzy inference process: for conjunction (i.e. ``and") the minimum operator is used; for disjunction (i.e. ``or") the maximum operator is used; for rule activation the minimum operator is used; for fuzzy rules aggregation the maximum operator is used. The fuzzy rules (see Table \ref{table:rule_base}) used in the control switcher were constructed using expert knowledge from HI data. In essence the fuzzy rules dictate that the EMICS will initiate a LOA switch only when the ``error'' is ``large'' and the robot is not reversing. If the ``error'' is ``large'' and the robot is reversing, the assumption is that the agent in control is trying to extricate the robot from error situation. Hence, by taking into account this simplified contextual knowledge, the control switcher will not switch LOA.

\begin{figure}
	\centering
	\includegraphics[width=0.5\columnwidth]{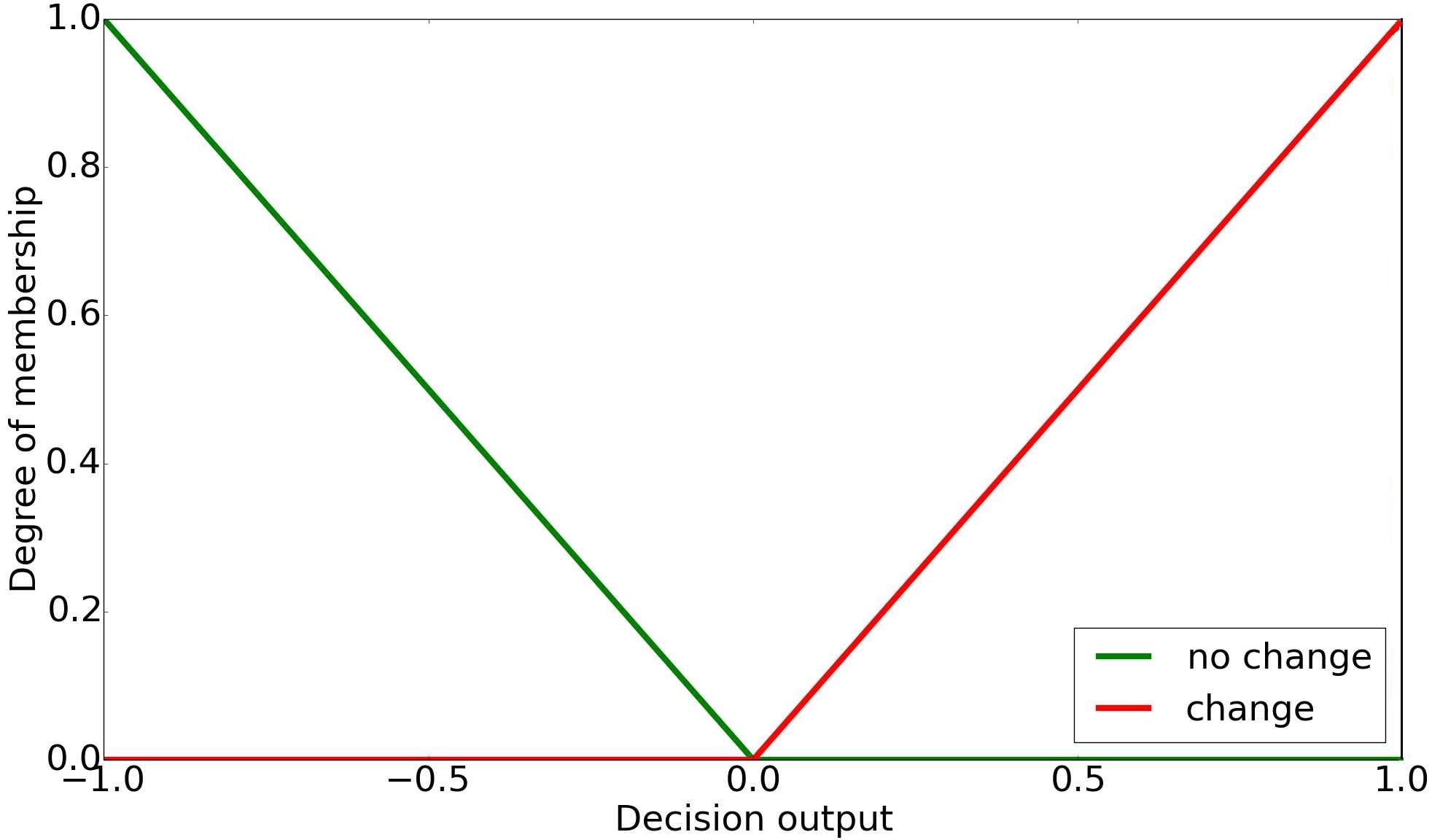}
	\caption{Output membership functions.} 
	\label{fig:fuzzy_output}
\end{figure}

Similar to the work of Nagi et al. \cite{Nagi2009} we follow the fuzzy bang-bang relay controller (FBBRC) approach. This means that the Largest of Maxima (LOM) defuzzification method is used. The output's universe of discourse $Y = [-1,1]$ represents the bang-bang output (see Figure \ref{fig:fuzzy_output}). The value of $y = -1$ means that no LOA switching initiative will take place. The value of $y = 1$ means that the control switcher will initiate a LOA switch.

The LOA switching problem under investigation is complex, ill-defined, and its underlying dynamics are not precisely known. Thus, a fuzzy control switcher offers several advantages. Primarily it allows the efficient management of the real world fuzziness through the use of expert knowledge coming from human operators. This can be easily achieved by extending the fuzzy sets and fuzzy rules base to include new linguistic input variables, metrics, heuristics etc. 

Additionally, the control switcher can be extended to have more output states in order to facilitate a more complex system. This is based on the fact that fuzzy logic allows for smoother transition between states. Lastly, fuzzy logic enables more transparency and better understanding of how the control switcher works, something important given the current state of the research.

\begin{table*}
\caption {The fuzzy rule base.}
    \begin{tabular}{|c|c|}
    \hline
    No. & Rules                                                                \\ \hline
    1   & \textbf{IF} error is small \textbf{OR} error is medium \textbf{THEN} LOA is no change \\ \hline
    2   & \textbf{IF} error is large \textbf{AND} speed is not reverse \textbf{THEN} LOA is change \\ \hline
    3   & \textbf{IF} speed is reverse \textbf{AND} error is large \textbf{THEN} LOA is no change  \\ \hline
    \end{tabular}
    \label{table:rule_base}
\end{table*}

 \begin{table*}
   \caption {The fuzzy membership functions for linguistic values of input variables ``error'' and ``speed'' (see Figure \ref{fig:fuzzy_input}). For ``error", trapezoid membership functions have been used. For ``speed'' membership functions, two trapezoid for ``reverse'' and ``forward'' and one triangular for ``zero", were used. The membership functions were heuristically chosen in order to smoothly overlap throughout the universe of discourse. This is a common practice when designing fuzzy controllers. The membership function for ``error large'' was chosen based on the error threshold calculated in Section \ref{threshold_controller}.}
     \begin{tabular}{|c|c|}
    
      \hline
      Linguistic value & Membership functions       \\ \hline
    
       error small   & $ \mu_{A_1^1}(x_1) = \begin{cases}
       0,  &  0.06 \leq x\\
       1  &  0  \leq x \leq 0.035  \\
       \frac{0.06-x}{0.06-0.035}  &  0.035  \leq x \leq 0.06 
      \end{cases} $ \\ \hline
    
     error medium   & $ \mu_{A_1^2}(x_1) = \begin{cases}
     0,  & x \leq  0.045, 0.08 \leq x\\
     \frac{x-0.045}{0.055-0.045}  &  0.045  \leq x \leq 0.055 \\
     1  &  0.055  \leq x \leq 0.065  \\
     \frac{0.08-x}{0.08-0.065}  &  0.065  \leq x \leq 0.08
     \end{cases} $ \\ \hline
  
     error large   & $ \mu_{A_1^3}(x_1) = \begin{cases}
     0,  & x \leq  0.065 \\
     \frac{x-0.065}{b-0.065}  &  0.065  \leq x \leq 0.085 \\
     1  &  0.085  \leq x \leq 0.1  \\
     \end{cases} $ \\ \hline
    
     speed reverse   & $ \mu_{A_2^1}(x_2) = \begin{cases}
     0,  & -0.02 \leq  x \\
     1  &  x \leq -0.03  \\
     \frac{-0.02-x}{-0.02-(-0.03)}  &  -0.03  \leq x \leq -0.02
     \end{cases} $ \\ \hline
    
     speed zero   & $ \mu_{A_2^2}(x_2) = \begin{cases}
     0,  & x \leq  -0.03, 0.03 \leq x\\
     \frac{x-(-0.03)}{0-(-0.03)}  &  -0.03  \leq x \leq 0 \\
     \frac{0.03-x}{0.03}  &  0 \leq x \leq 0.03
     \end{cases} $ \\ \hline
  
     speed forward   & $ \mu_{A_2^3}(x_2) = \begin{cases}
     0,  & 0.02 \leq  x \\
     \frac{x-0.02}{0.03-0.02}  &  0.02  \leq x \leq 0.03 \\
     1  &  0.03  \leq x   \\
     \end{cases} $ \\ \hline

     \end{tabular}
     \label{table:membership_functions}
 \end{table*}
 
\subsection{Relevance of the Mixed-Initiative control switcher to shared control}
\label{section:shared_control_relevance}

As explained in Section \ref{lit_review_MI-systems}, we consider shared control to be a specific LOA that only loosely falls under the banner of MI control. This is despite the similarities that shared control and input blending systems have with MI control. This is because in most cases of shared control only the robot's AI controller has the initiative to adjust operator's control input e.g. to blend robot commands with operator's joypad navigation commands. In this section we will discuss how our MI control approach relates to shared control by using the formalism proposed by Dragan and Srinivasa \cite{Dragan2013}.  In their work they interpret shared control as a policy blending. More specifically, ``as an arbitration of two policies, namely, the user’s input and the robot’s prediction of the user’s intent. At any instant, given the input, $U$, and the prediction, $P$, the robot combines them using a state-dependent arbitration function $a \in [0, 1]$''. 

Let us assume a navigation task similar to our case. Let us also assume that the navigation goal or a correct prediction $P$ of that goal, is known. Then, according to the above formalism, the arbitration function $a$ will decide on how the robot's AI control input will be blended with the operator's control commands $U$ to contribute towards navigating to the goal. Simply put, the more the operator's commands are in accordance to the goal or prediction $P$, the less the AI controller will blend its commands with $U$. If $U$ is not in accordance with $P$ then the AI controller will heavily adjust $U$ so that the output commands will contribute to navigation towards the goal. In this example two extreme cases exist: a) the operator's input $U$ does not need any correction and hence it is used as the output command directly; or b) the AI controller fully blends/changes $U$ and hence the AI controller's commands are used as output. These two extreme cases roughly correspond to our MI approach with the equivalent of the arbitration function being the EMICS. Case a) corresponds to the teleoperation LOA (operator fully in control); and case b) corresponds to the autonomy LOA (robot's autonomous navigation capabilities fully in control). 

However, there are fundamental differences between the above formalism and our EMICS. In shared control the arbitration function decides how the operator's input will be blended with the robot's AI commands in a continuous range from teleoperation to autonomy, and only in a very specific control element. This control element usually lies in low level control, e.g. velocity navigation commands. Also the initiative lies within the arbitration function. In contrast, the EMICS proposed here allows switches between different discrete LOAs with both the operator and the EMICS having the authority to initiate or override each other's actions. 

The framework and the control switcher presented in this paper (see Figure \ref{fig:block_diagram}) are applicable and can be extended to also include multiple LOAs without changing their fundamental principles. For example a third LOA such as shared-control can be added. One way to do this is by extending the EMICS's rule base and the output membership function. A rule can be added to switch to shared control LOA when the ``error'' is ``medium''. The practical interpretation of such LOA switch is that the operator needs some assistance but not necessarily in the form of the AI controller taking full control. Similarly, additional LOAs can be added depending on the application, e.g. a LOA in which the robot will perform a certain function (e.g. navigate autonomously) while the operator will perform some other function concurrently (e.g. control of a pan-tilt camera or of robot's arm). However, because the problem of MI switching between two LOAs is already challenging, we decided to perform our investigation by implementing the two LOAs corresponding to the extreme cases: teleoperation and autonomous control. 

\section{Experiment 1: evaluation using a simulated robot and test environment}
\label{section:experiment2_2}

To allow a direct comparisons between MI and HI control we first evaluate our EMICS using the same framework as in our previous work \cite{Chiou2016}. To be useful, the MI algorithm should provide the same level of performance or better in terms of primary task completion time or score when compared to the HI system. The reason that the MI control switcher can be useful despite potentially having the same level of performance as HI is that there are situations in which a human-initiated LOA switch may not be possible (e.g. loss of communication or incapacitated operator), therefore the MI control switcher has the ability to provide additional system redundancy which is not evaluated in these experiments.

An experimental evaluation of the EMICS described in Section \ref{fuzzy_controller} was conducted. The aim was to make an initial evaluation of the EMICS and compare it's performance with that of the HI control switcher. If performance on the tasks prove to be in a similar level or better than the HI control switcher, then the EMICS has a positive and meaningful impact upon the system. This will be especially true compared to using only teleoperation or autonomy. More specifically the experiment described in this section evaluates: a) the human's and EMICS's ability to switch LOA between teleoperation and autonomy in order to overcome circumstances in which an MI system is under-performing; b) how the EMICS performs compared to the HI one of \cite{Chiou2016}; c) the unfolding Human-Robot Interaction (HRI) of MI control. 

\subsection{Experimental setup - test arena, tasks and tested conditions}

\begin{figure*}
	\centering
	\begin{subfigure}[b]{0.4\textwidth}
		\centering
		\includegraphics[width=\textwidth]{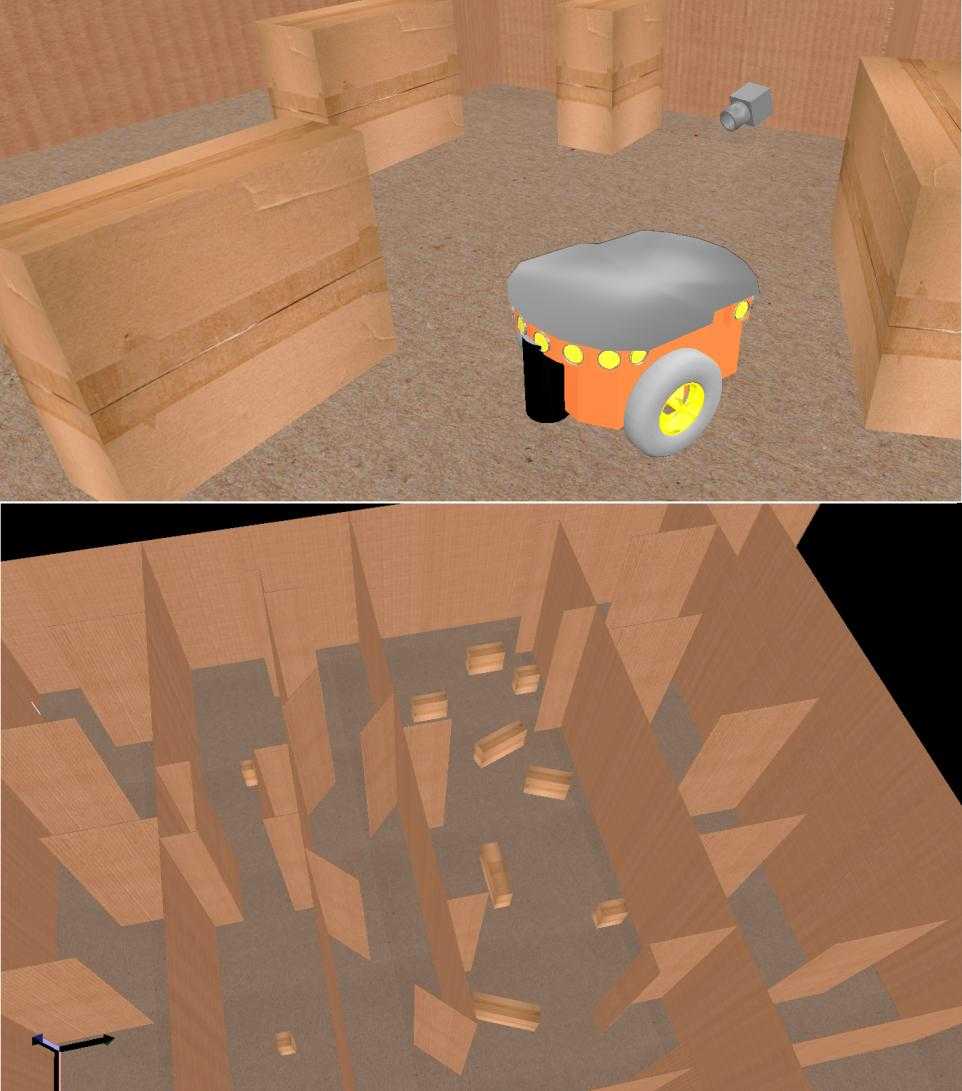}
		\caption{}
		\label{subfig:arena_exp2}
	\end{subfigure}
	\hfill
	\begin{subfigure}[b]{0.4\textwidth}
		\centering
		\includegraphics[width=\textwidth]{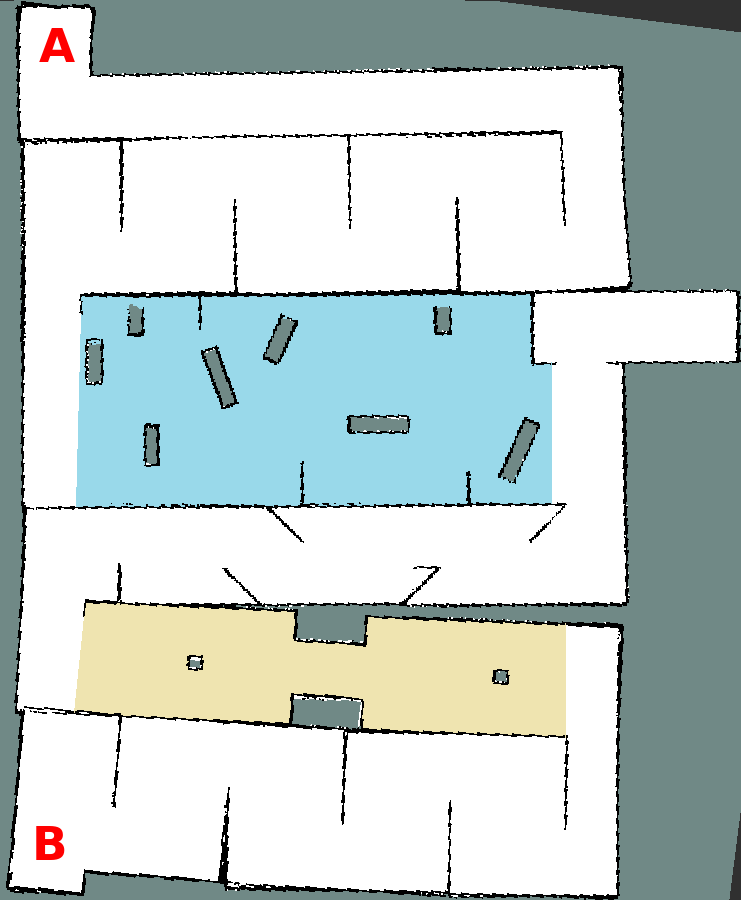}
		\caption{}
		\label{subfig:map_exp2}
	\end{subfigure}
	\hfill
	\caption{\textbf{\ref{subfig:arena_exp2}:} The simulated arena and the robot model used in the experiment. \textbf{\ref{subfig:map_exp2}:} laser-derived SLAM map created in the simulation environment. Primary task was to drive from point A to B and back again to A. The yellow shaded region is where artificial sensor noise was introduced. The blue shaded region is where the secondary task was presented to the operator.}
	\label{fig:arenas_exp2}
\end{figure*}
    
In the previous work \cite{Chiou2016} we carried out experiments in a high fidelity simulated arena with dimensions of $11 \times 13.5$ meters (see Figure \ref{fig:arenas_exp2}) using Human-Initiative (HI) control. The robot was controlled by an OCU, composed of a laptop, a joypad, a mouse and a screen showing the control interface. We used a simulated Pioneer-3DX mobile robot fitted with a camera and a laser scanner. We conducted the experiment described in this section using an identical setup to \cite{Chiou2016}, i.e. identical robot arena; system; experimental paradigm and procedures. This is in order to facilitate direct comparison between the control switchers. 

Operators were given the primary task of navigating from point A in Figure \ref{subfig:map_exp2} (the beginning of the arena) to point B (the end of the arena) and then back again to point A. During each experimental trial, two different kinds of performance-degrading factors were introduced, one for each agent. At certain times, artificially generated sensor noise was used to degrade the performance of autonomous navigation. At other times, a cognitively intensive secondary task was used to degrade the performance of the human operator. Each of these performance-degrading situations occurred twice per experimental trial, once on the way from point A to point B, and once on the way back from B to A. These degrading factors occurred separately from each other, as shown in Figure \ref{subfig:map_exp2}.

More specifically, autonomous navigation was degraded by adding Gaussian noise to the laser scanner range measurements, thereby degrading the robot’s AI localization and obstacle avoidance abilities. To ensure experimental repeatability, this additional noise was instantiated whenever the robot entered a predefined area of the arena, and was deactivated when the robot exited that area. Note that such region-specific noise can in fact happen in real-world applications, e.g. when a robot travels through a highly radioactive region during nuclear decommissioning or the exploration and remediation of nuclear disaster sites such as Fukushima. 

\begin{figure*}
		\centering
		\begin{subfigure}[b]{0.4\textwidth}
			\centering
			\includegraphics[width=\textwidth]{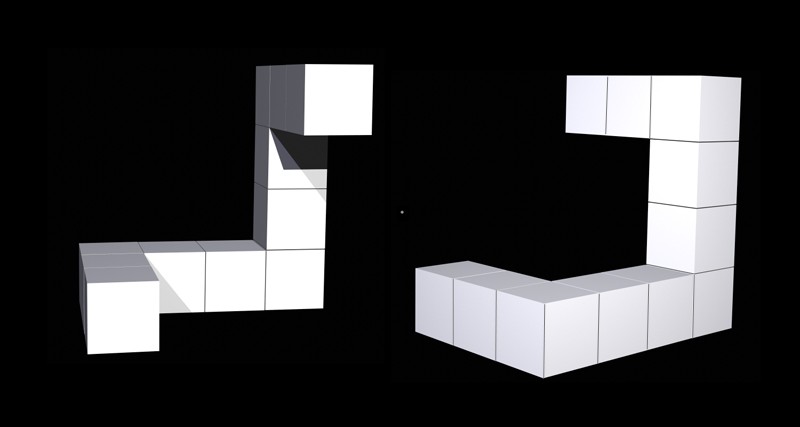}
			\caption{}
			\label{fig:rotation_task_exp2}
		\end{subfigure}
		\hfill
		\begin{subfigure}[b]{0.5\textwidth}
			\centering
			\includegraphics[width=\textwidth]{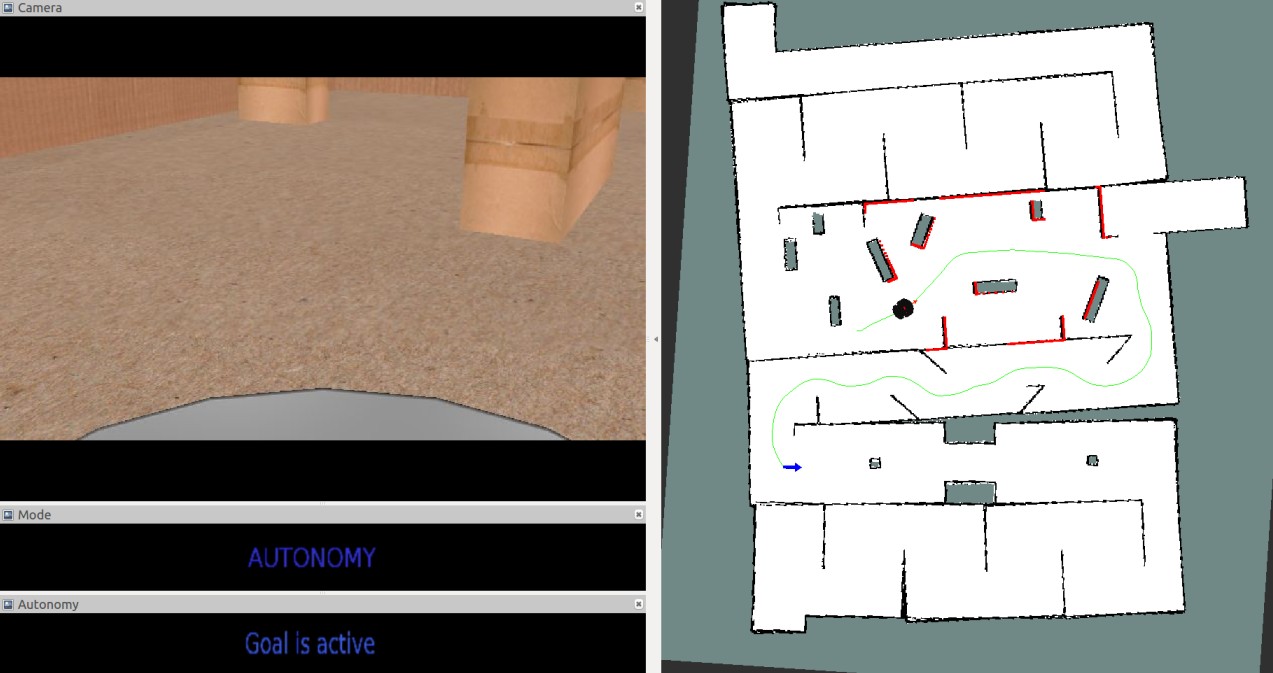}
			\caption{}
			\label{subfig:interface_sim}
		\end{subfigure}
		\hfill
		\caption{\textbf{\ref{fig:rotation_task_exp2}:} A typical example of a rotated 3D objects card. \textbf{\ref{subfig:interface_sim}:} The control interface as presented to the operator. \textbf{Left}: video feed from the camera, the control mode in use and the status of the robot. \textbf{Right}: The map showing the position of the robot, the current goal (blue arrow), the navigation planner path (green line), the obstacles' laser reflections (red) and the walls (black).}
		\label{fig:interface_rotation}
	\end{figure*}

To degrade the performance of the human, a secondary task of mentally rotating 3D objects was used \cite{Ganis2015}. Whenever the robot entered a predefined area in the arena, the operator was presented with a series of 10 cards, each showing images of two 3D objects (see Figure \ref{fig:rotation_task_exp2}). These cards were put by the experimenter on the OCU desk, on the right hand side of the operator. On five cards, the objects were identical but rotated, and on the other five cards the objects were mirror images with opposite chiralities. The operator was required to state whether or not the two objects were identical.

In \cite{Chiou2016} for each human test-subject, three different control modes were tested: a) \textbf{teleoperation} mode, in which the human operator was restricted to using only direct joypad control to drive the robot; b) \textbf{autonomy} mode, in which the operator was only allowed to guide the robot by clicking desired destinations on the robot’s laser-generated 2D map; c) \textbf{HI} mode, in which the operator was allowed to switch LOA at any time (using a button on the joypad) according to their judgement, in order to maximize performance.

In the experiment presented here each operator was tested in \textbf{Robot-Initiative (RI)} and \textbf{MI} control. The EMICS, as described in Section \ref{fuzzy_controller}, gives the MI system and the operator the capacity and authority to switch dynamically (i.e. during task execution) between teleoperation and autonomy. The EMICS initiates LOA switches based on \textit{goal directed motion} performance (i.e. effectiveness) and simplified (i.e. limited) context. The operator switches LOA using a button on the joypad, according to their judgement. The RI control switcher uses identical mechanisms with the EMICS for the switching behavior. However, it removes the operator's ability to trigger LOA switches and thus it restricts them to using the RI control switcher's dictated LOA. When a LOA switch occurs, the control switchers alert operators in three different ways using: a) an alarm sound identical to the one denoting ``engine failure'' in aircrafts; b) synthetic speech expressing the LOA the control switcher have switched to; c) a GUI notification.

\subsection{Participants and procedure}
The 24 participants of our previous HI study were asked to participate in this experiment. From these, 16 were available at the time of the experiment and participated. This allowed us, along with the identical setup, to directly compare the results from MI and RI with the ones of the same participants from the previous experiment in order to minimize between-participants variation. The majority of participants were recruited from the university's student population with the majority been males in their early 20s. We used a within-groups experimental design with every participant performing two trials: a) one using the EMICS; b) one using the RI control switcher. 

Each participant underwent extensive standardized training before the experiment, similar to our previous work \cite{Chiou2016, Chiou2019_SMC_learning_effects}. Each system aspect was introduced gradually and participants were allowed to try them in practice. Additionally, participants were not allowed to proceed with the experimental trials until they had first demonstrated that they could complete a training obstacle course within a specific time limit and with no collisions. This ensured that all participants had adequate understanding of the new system and had attained a common minimum skill level. Counterbalancing was used in the experimental trials (i.e. the order of the tested control switchers was rotated for different participants) in order to prevent both learning and fatigue effects from biasing the results. 

Participants were asked to perform the primary task (robot navigation) as quickly as possible while minimizing collisions. Participants were also instructed that, when presented with the secondary task, they should do it as quickly and as accurately as possible. They were explicitly told to prioritize the secondary task over the primary task, and only to perform the primary task if the workload allowed. This is to prevent different operators having different priorities (e.g. some of them focusing on driving while others focusing on the secondary task) and thus minimizing a potential confounding factor. Additionally, as explained in our previous work \cite{Chiou2016}, when people are instructed to do both tasks in parallel to the best of their abilities, they either a) ignore the secondary task or b) choose random answers for the secondary task to alleviate themselves from the secondary workload, so that they can continue focusing on the primary task of robot driving.

The operators could only acquire SA information via the OCU, which displays real-time video feed from the robot’s front-facing camera, and the estimated robot location on the 2D SLAM map (for the interface see Figure \ref{subfig:interface_sim}). All participants were given an identical and complete 2D map, generated by SLAM prior to the trials.

At the end of each experimental run participants completed a NASA Task Load Index (TLX) questionnaire \cite{Sharek2011}. NASA-TLX is a questionnaire system which enables the perceived workload and difficulty of using a system to be numerically quantified.

\subsection{Results: tasks performance}

\begin{table*}[ht]
\caption{Table showing the ANOVA results and the descriptive statistics for the metrics used in the experiment.}
	\centering
	\begin{tabular}{lll}
		\hline
		\textbf{metric}                                                                 & \textbf{ANOVA control mode effect}                                                                               & \textbf{descriptive statistics}                                                                                                                                                        \\ \hline
		\begin{tabular}[c]{@{}l@{}}primary task\\ completion time\end{tabular}          & \begin{tabular}[c]{@{}l@{}}$F(2, 30) = 19.116$, $p < .01$,\\ $power = 1$, $\eta^2 = .56$\end{tabular}            & \begin{tabular}[c]{@{}l@{}}HI: $M = 418.5$ $sec$, $SD = 34.9$\\ MI: $M = 374.2$ $sec$, $SD = 15.6$ \\ RI: $M = 392.3$ $sec$, $SD = 22.4$\end{tabular}                                  \\ \hline
		collisions                                                                      & \begin{tabular}[c]{@{}l@{}}$F(1.469, 22.036) = 1$, $p > .05$, \\ $power < 0.8$, $\eta^2 = .062$\end{tabular}     & \begin{tabular}[c]{@{}l@{}}HI: $M = .63$, $SD = 1.1$ \\ MI: $M = 0.38$, $SD = 0.6$ \\ RI: $M = .81$, $SD = 1.5$\end{tabular}                                                           \\ \hline
		\begin{tabular}[c]{@{}l@{}}primary task \\ score\end{tabular}                   & \begin{tabular}[c]{@{}l@{}}$F(2, 30) = 16.774$, $p < .01$, \\ $power > 0.9$, $\eta^2 = .528$\end{tabular}        & \begin{tabular}[c]{@{}l@{}}HI: $M = 424.8$, $SD = 36.2$ \\ MI: $M = 378$, $SD = 18.7$ \\ RI: $M = 400.4$, $SD = 22.9$\end{tabular}                                                     \\ \hline
		\begin{tabular}[c]{@{}l@{}}secondary task \\ completion time\end{tabular}       & \begin{tabular}[c]{@{}l@{}}$F(3, 45) = 10.344$, $p < .01$, \\ $power > .95$, $\eta^2 = .408$\end{tabular}        & \begin{tabular}[c]{@{}l@{}}HI: $M = 40$ $sec$, $SD = 11.1$ \\ MI: $M = 34.5$ $sec$, $SD = 10.4$ \\ RI: $M = 33.2$ $sec$, $SD = 8.6$ \\ baseline: $M = 34$ $sec$, $SD = 6$\end{tabular} \\ \hline
		\begin{tabular}[c]{@{}l@{}}secondary task \\ errors\end{tabular}                & \begin{tabular}[c]{@{}l@{}}$F(3, 45) = 0.891$, $p > .05$, \\ $power < .8$, $\eta^2 = .056$\end{tabular}          & \begin{tabular}[c]{@{}l@{}}HI: $M = 1.7$, $SD = 1.67$ \\ MI: $M = 1.4$, $SD = 1.4$ \\ RI: $M = 2$, $SD = 1.67$ \\ baseline: $M = 1.7$, $SD = 1.7$\end{tabular}                         \\ \hline
		\begin{tabular}[c]{@{}l@{}}NASA-TLX\\ scores\end{tabular}                       & \begin{tabular}[c]{@{}l@{}}$F(1.337, 20.049) = 10.135$, $p < .01$, \\ $power > .9$, $\eta^2 = .403$\end{tabular} & \begin{tabular}[c]{@{}l@{}}HI: $M = 43.561$, $SD = 13.026$ \\ MI: $M = 33.217$, $SD = 13.399$ \\ RI: $M = 28.899$, $SD = 9.642$\end{tabular}                                           \\ \hline
		\begin{tabular}[c]{@{}l@{}}percentage of time \\ spent in autonomy\end{tabular} & \begin{tabular}[c]{@{}l@{}}$F(2, 25) = 0.74$, $p > .05$,\\ $power < .8$, $\eta^2 = .50$\end{tabular}             & \begin{tabular}[c]{@{}l@{}}HI: $M = 57.7\%$, $SD = 25.9$ \\ MI: $M = 53\%$, $SD = 23.25$ \\ RI: $M = 49.1\%$, $SD = 12.22$\end{tabular}                                                \\ \hline
		\begin{tabular}[c]{@{}l@{}}number of \\ LOA switches\end{tabular}               & \begin{tabular}[c]{@{}l@{}}$F(2, 30) = 16.78$, $p < .01$, \\ $power > .9$, $\eta^2 = .528$\end{tabular}          & \begin{tabular}[c]{@{}l@{}}HI: $M = 10.4$, $SD = 8.64$ \\ MI: $M = 13.6$, $SD = 11.78$ \\ RI: $M = 3.13$, $SD = 1.15$\end{tabular}                                                     \\ \hline
	\end{tabular}
	\label{table:results_exp2_2}
\end{table*}

A repeated measures one-way ANOVA was used to compare RI, MI and HI. An ANOVA with Greenhouse-Geisser correction was used for the cases where the sphericity assumption (i.e. that the variances of the differences between trials are not equal) was violated. For HI, the subset of data from \cite{Chiou2016} corresponding to the 16 participants was used. Although a large period of time (7 months) had passed since their initial participation, we identified learning as a factor that might have affected performance (see Section \ref{section:learning_discussion}). Fisher's least significant difference (LSD) test was used for pairwise comparisons. Post-hoc adjustments such as Bonferroni were not used in this paper given the a) clear hypothesis; b) predefined post-hoc comparisons; c) small number of comparisons; d) early stage of research in this domain (i.e. to avoid missing important findings or interesting insights due to inflated type II error \cite{Streiner2011,Rothman1990,Perneger1998}).

We consider a result to be significant when it yields a $p$ value less than $0.05$. Lastly we report on the statistical power of the results and the effect size. The detailed statistical calculations are reported in Table \ref{table:results_exp2_2}. In all graphs throughout the paper the error bars indicate the standard error and the mean, the dots the data points, and the violin plots the smoothed probability density of the data. Also, in the graphs, the results of the pairwise comparisons are plotted when significant or relevant with a * for $p<0.05$, ** for $p<0.01$, *** for $p<0.001$, and with $ns$ when no significant result was found.

\begin{figure*}
		\centering
		\begin{subfigure}[b]{0.49\textwidth}
			\centering
			\includegraphics[width=\textwidth]{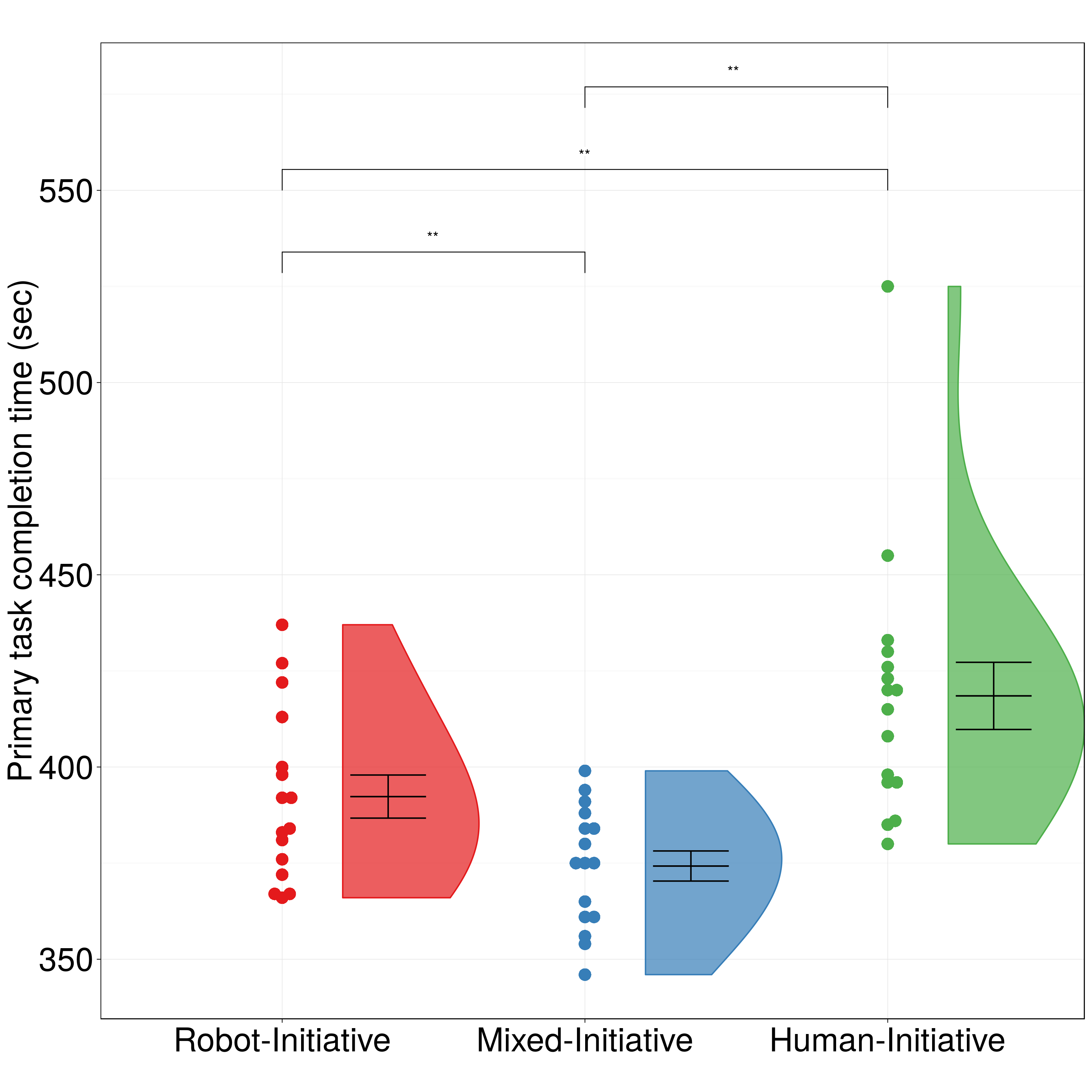}
			\caption{}
			\label{subfig:primary_time_exp2_2}
		\end{subfigure}
		\hfill
		\begin{subfigure}[b]{0.49\textwidth}
			\centering
			\includegraphics[width=\textwidth]{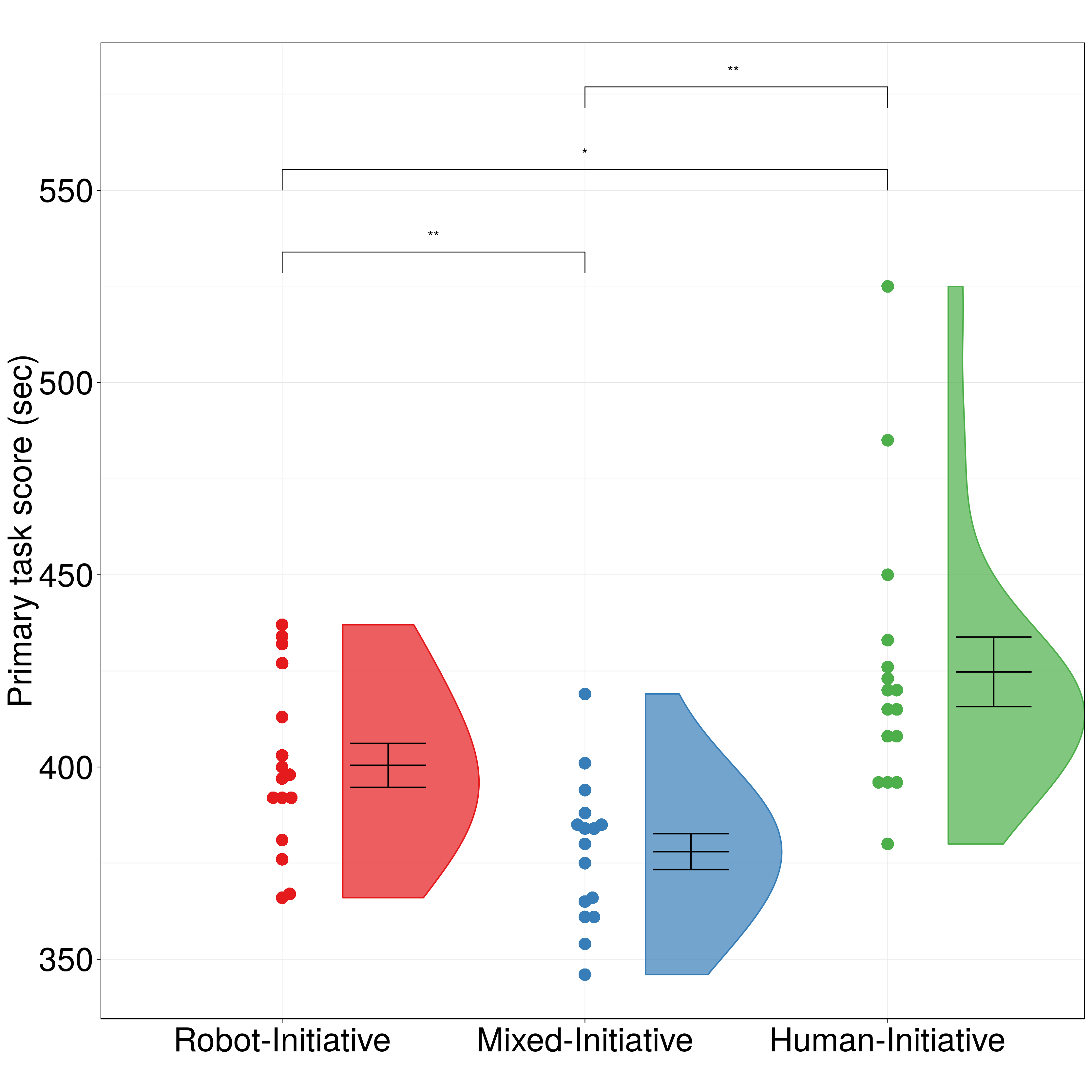}
			\caption{}
			\label{subfig:primary_score_exp2_2}
		\end{subfigure}
		\hfill
		\caption{\textbf{\ref{subfig:primary_time_exp2_2}:} Primary task time-to-completion. \textbf{\ref{subfig:primary_score_exp2_2}:} Primary task score combining time and collisions penalty. MI outperformed significantly both HI and RI in the primary task.}
		\label{fig:primary_exp2_2}
	\end{figure*}

ANOVA for \textit{primary task completion time} (see Figure \ref{subfig:primary_time_exp2_2}) showed overall significantly different means between HI variable-autonomy, RI and MI. Pairwise comparison revealed that HI performed significantly worse (i.e. slower completion time) than the other two control modes with \textit{$p <.01$}. Also MI variable autonomy performed significantly better than RI (\textit{$p <.01$}). 

The effect of control mode on the number of \textit{collisions} was not significant between RI, MI and HI variable autonomy mode.

We used a \textit{primary task score} in order to be able to capture the speed-accuracy trade-off that different operators might have (i.e. how fast an operator is driving the robot vs how carefully). The primary task score (see Figure \ref{subfig:primary_score_exp2_2}) is calculated by adding a time penalty of \textit{$10$ $sec$} for every collision, onto the primary task completion time for each participant. This is inspired by the performance scores used in the RoboCup competitions \cite{Jacoff2003a}. ANOVA analysis showed that control mode had a significant effect on primary task score. The LSD tests suggests that HI variable autonomy is significantly (\textit{$p <.01$}) worse than the MI mode and the RI mode ( \textit{$p =.01$}). Lastly, MI significantly (\textit{$p <.01$}) outperformed RI regarding primary task score. 

\begin{figure*}
		\centering
		\begin{subfigure}[b]{0.49\textwidth}
			\centering
			\includegraphics[width=\textwidth]{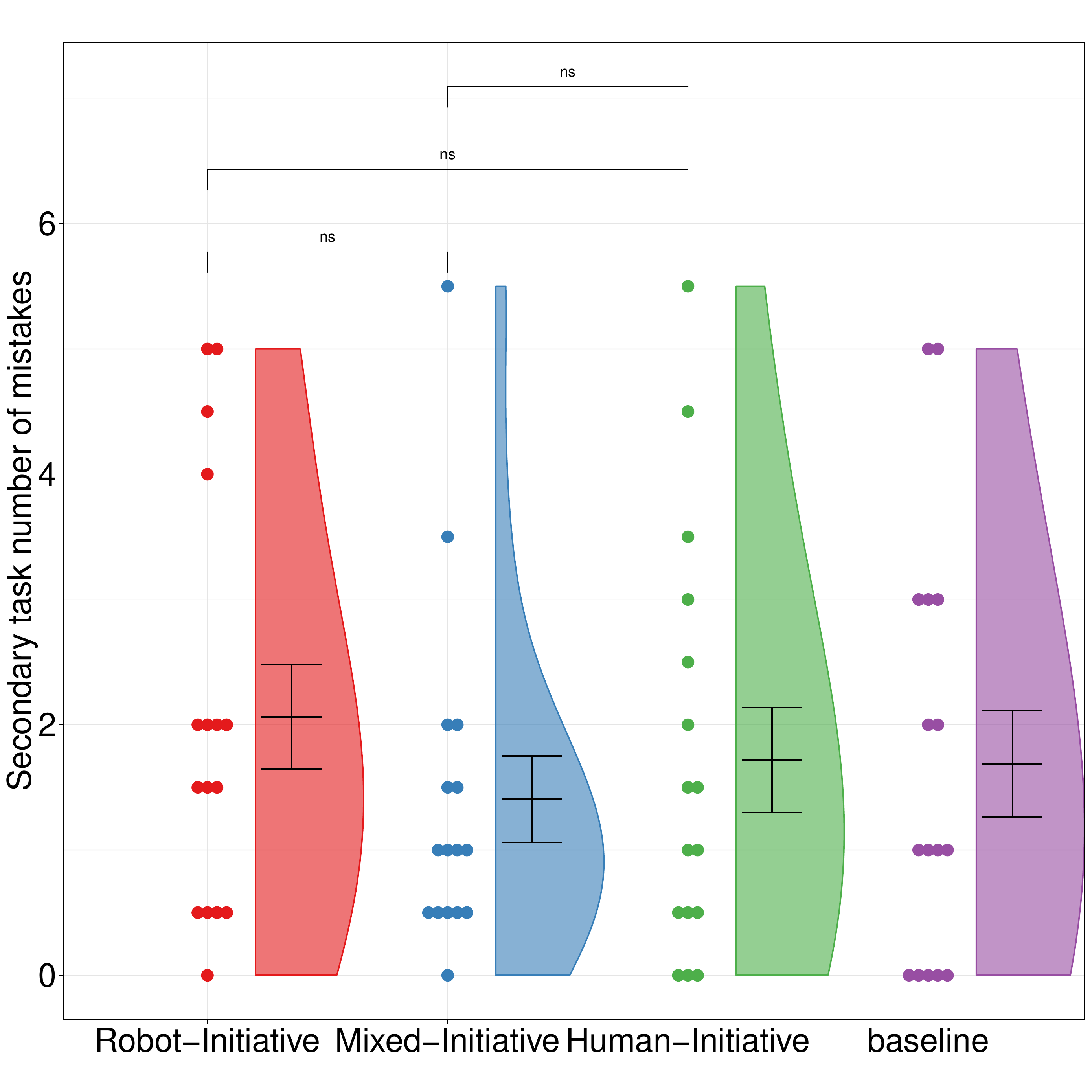}
			\caption{}
			\label{subfig:secondary_mistakes_exp2_2}
		\end{subfigure}
		\hfill
		\begin{subfigure}[b]{0.49\textwidth}
			\centering
			\includegraphics[width=\textwidth]{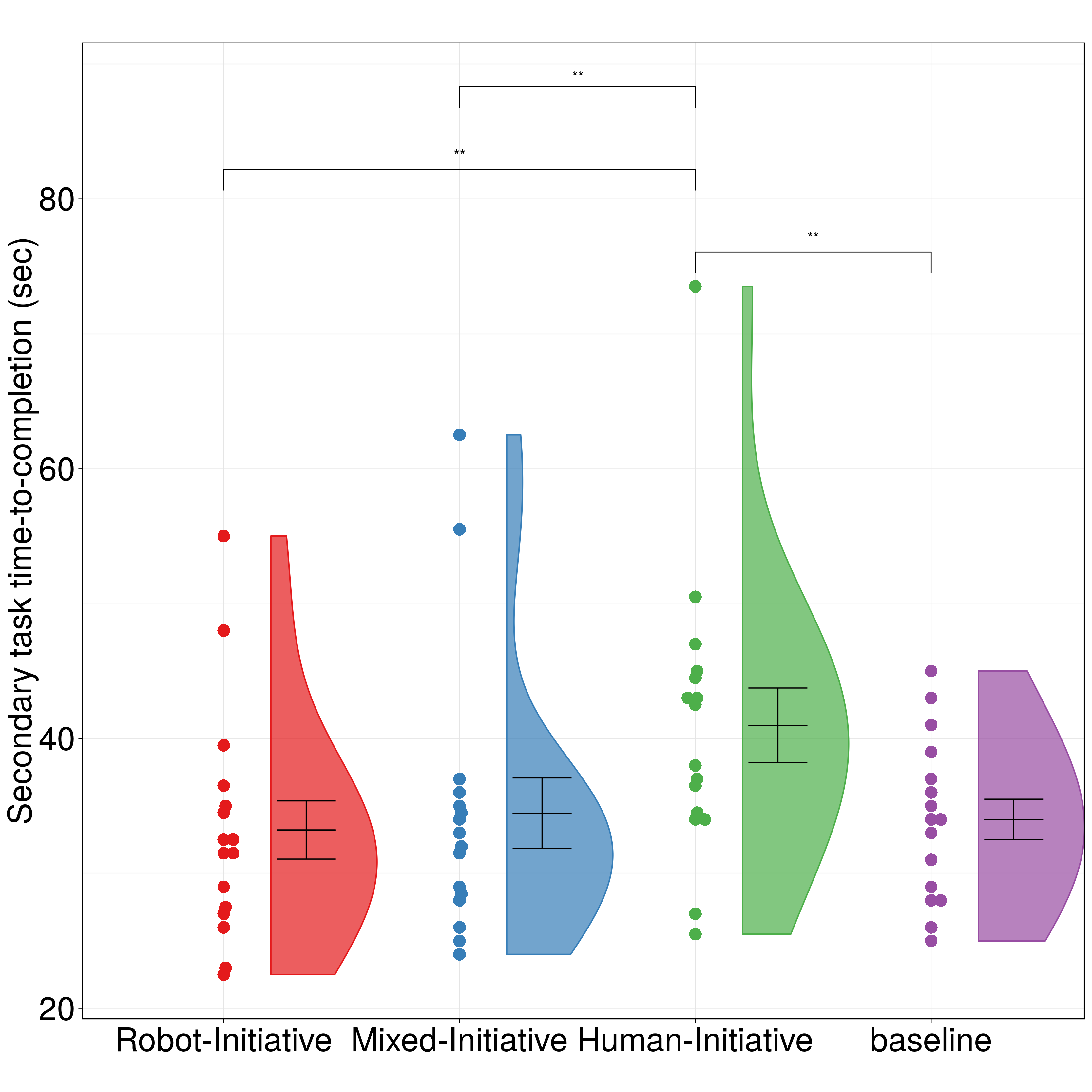}
			\caption{}
			\label{subfig:secondarytime_exp2_2}
		\end{subfigure}
		\hfill
		\caption{\textbf{\ref{subfig:secondary_mistakes_exp2_2}:} Secondary task number of mistakes/errors. \textbf{\ref{subfig:secondarytime_exp2_2}:} Secondary task completion time.}
		\label{fig:secondary_exp2_2}
	\end{figure*}

The average time per trial that the participants took to complete one series of the 3D object rotations, is denoted by the \textit{Secondary task completion time} (see Figure \ref{subfig:secondarytime_exp2_2}). ANOVA showed that there is a significant difference between the means of secondary task completion times. HI variable autonomy mode performed significantly worse (i.e. more time to complete) than the other modes with $p < .01$. Performance between the baseline trial, the MI mode and RI mode were without any significant difference ($p > .05$, i.e. same level of performance).

The \textit{number of secondary task errors} (see Figure \ref{subfig:secondary_mistakes_exp2_2}) is the average number of mistake/errors per trial that the participants made during one series of the 3D object rotations. The number of secondary task errors did not have significant differences between the different control modes as shown by ANOVA.

Control mode had a significant effect on \textit{NASA-TLX scores} (see Figure \ref{subfig:NASA-TLX_exp2_2}) as shown by ANOVA. Pairwise comparisons showed that RI was perceived by participants as having the lowest difficulty, as compared to HI variable autonomy mode with $p < 0.01$ and MI mode with $p = 0.05$. HI variable autonomy is perceived as being more difficult than MI ($p < 0.01$). 

\subsection{Results: Human-Robot-Interaction}
\label{section:results_HRI_sim}
Similar to our previous work \cite{Chiou2016_AAAI} we analyzed the average \textit{percentage of time spent in autonomy} for each controller. Due to corrupted data for one of the participants, data from 15 participants was used for every result reported on this particular metric. This is contrary to the rest of the analysis in which data from all 16 participants was used. ANOVA did not show any effect caused by control mode in the average percentage of time spent in autonomy.

\begin{figure*}
		\centering
		\begin{subfigure}[b]{0.49\textwidth}
			\centering
			\includegraphics[width=\textwidth]{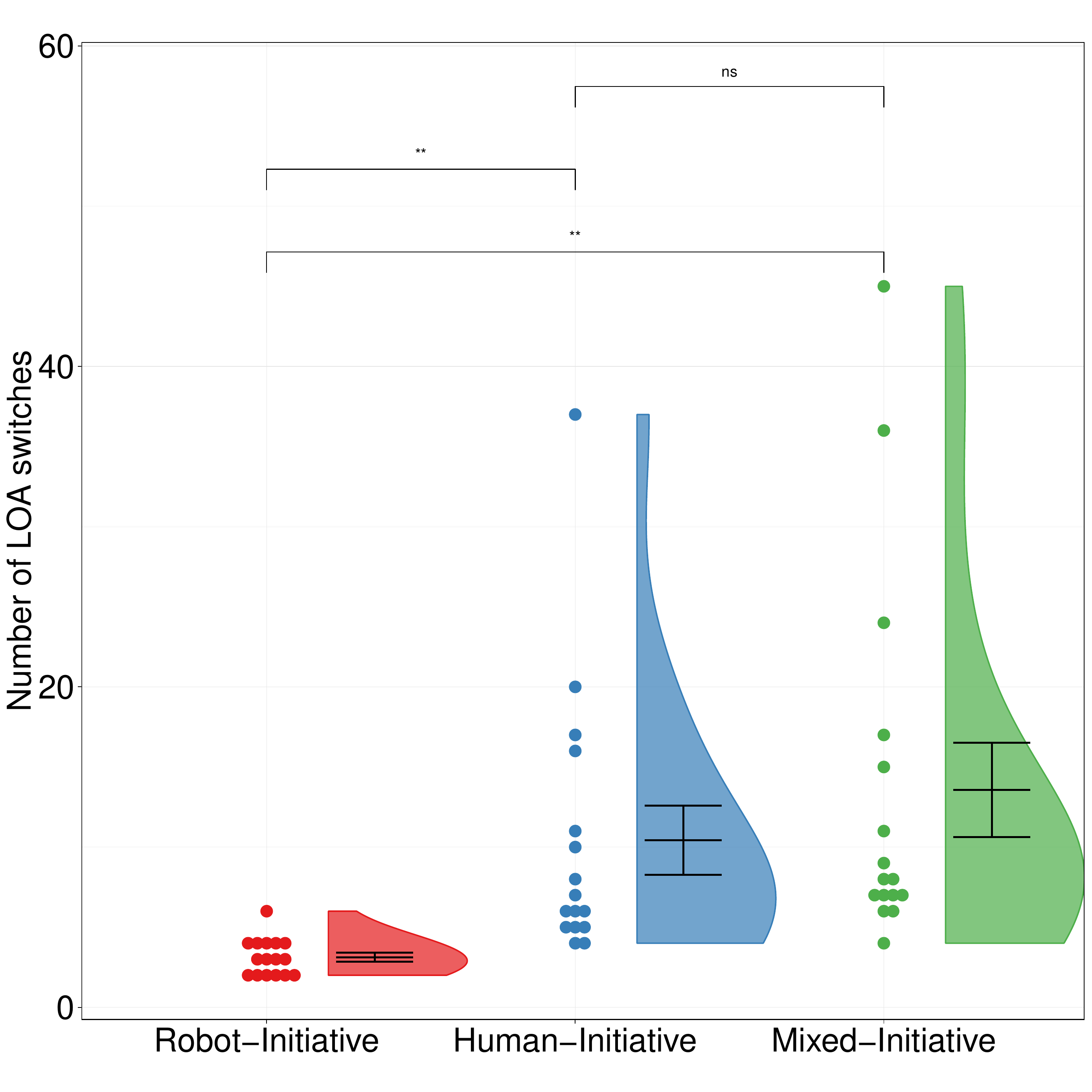}
			\caption{}
			\label{fig:number-loa-switches_exp2_2}
		\end{subfigure}
		\hfill
		\begin{subfigure}[b]{0.49\textwidth}
			\centering
			\includegraphics[width=\textwidth]{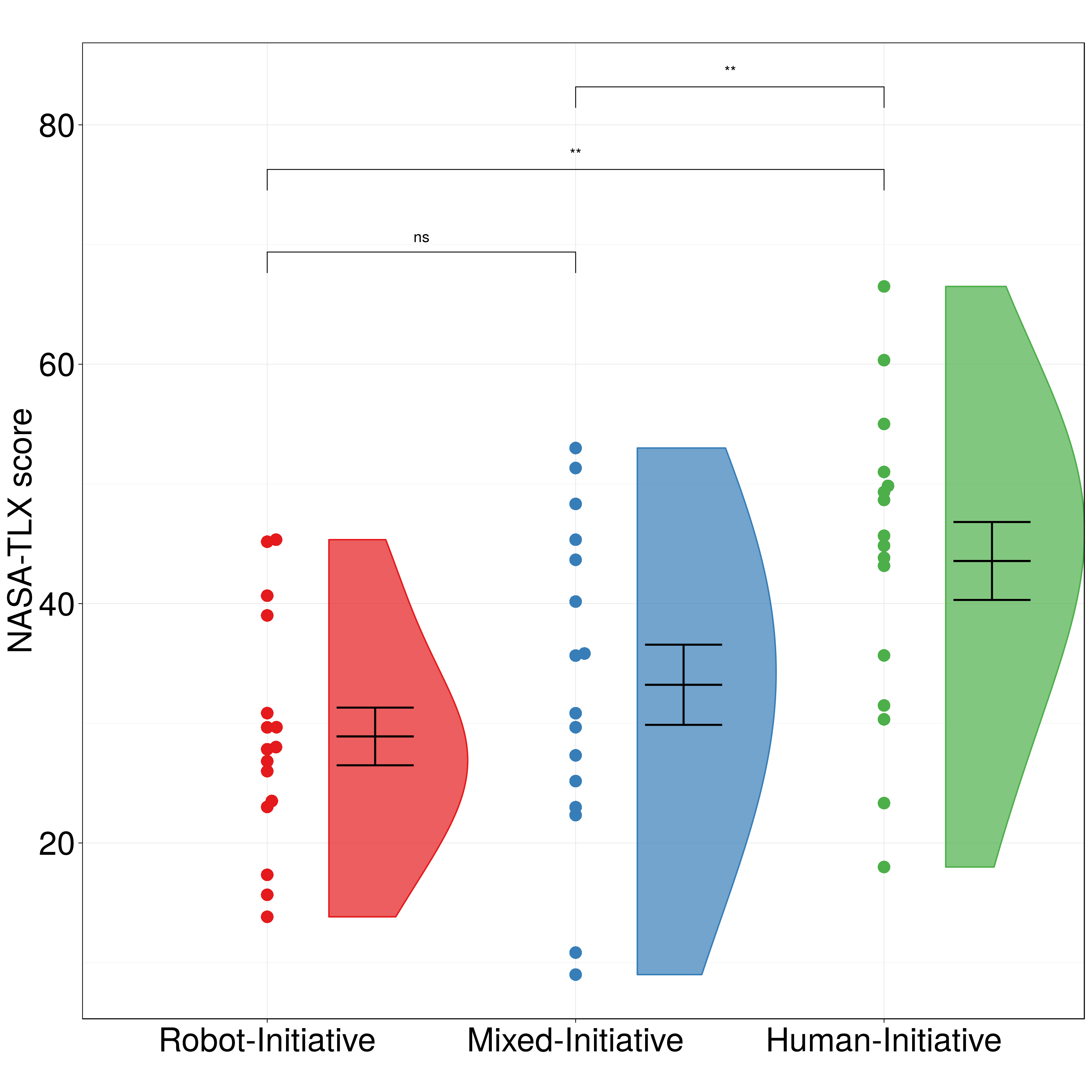}
			\caption{}
			\label{subfig:NASA-TLX_exp2_2}
		\end{subfigure}
		\hfill
		\caption{\textbf{\ref{fig:number-loa-switches_exp2_2}:} The number of LOA switches per control mode. \textbf{\ref{subfig:NASA-TLX_exp2_2}:} NASA-TLX score showing the overall trial difficulty/workload as perceived by the operators.}
		\label{fig:loa_tlx}
	\end{figure*}

The \textit{number of LOA switches} (see Figure \ref{fig:number-loa-switches_exp2_2}) performed in each trial denotes the frequency in which operators switched LOA in HI; frequency that operators and the EMICS switched LOA in MI; and frequency in which the RI control switcher (i.e. the restricted EMICS) switched LOA in RI. ANOVA showed that control mode had a significant effect on the number of LOA switches. Pairwise comparisons showed that RI had significantly ($p < 0.01$) fewer LOA switches as compared to HI mode and MI mode. The number of LOA switches of HI and MI control modes were on the same level (i.e. no statistical difference, $p > 0.05$). Out of the $13.6$ MI LOA switches, $1.3$ were initiated from the EMICS.

We performed a correlation analysis using a two-tailed Pearson's $r$ to investigate any relationships between different metrics and other variables. The \textit{number of LOA switches} in HI and the \textit{number of LOA switches} in MI are highly correlated ($r(14)= .849, p<.01 $). Similar to HI (see \cite{Chiou2016_AAAI}) there was no correlation between the \textit{number of LOA switches} and performance in the \textit{primary task score} or \textit{secondary task completion time} in RI and in MI.

No correlation was found between the \textit{percentage of time spent in autonomy} in HI and MI ($r(13)= .202, p>.05$); and between the \textit{percentage of time spent in autonomy} in HI and RI ($r(13)= .094, p>.05$). No correlation was found between MI and RI \textit{percentage of time spent in autonomy} ($r(13)= .326, p>.05$).

The \textit{percentage of time spent in autonomy} LOA is positively correlated with the \textit{primary task score} in RI ($r(13)= .528, p<.05$). Also the same metrics are positively correlated in MI ($r(13)= .627, p<.05$). As expected given results in \cite{Chiou2016_AAAI}, no correlation was found between \textit{primary task score} in HI and \textit{percentage of time spent in autonomy} in HI ($r(13)= .047, p>.05$).

Lastly, no correlation was found between \textit{time spent in autonomy} and \textit{secondary task completion time} for any of the 3 control modes, HI (as found in \cite{Chiou2016_AAAI}), MI, and RI.

\subsection{Discussion}
\label{section:learning_discussion}
Mixed-Initiative (MI) and Robot-Initiative (RI) outperformed Human-Initiative (HI) in terms of primary task performance. What this shows primarily, in the context of this experiment, is that the fuzzy robot control switcher (i.e. both in MI and in RI) is capable of successfully measuring human-robot system performance, inferring if a switch in the LOA is needed, and initiating a LOA switch. This is particularly true given that RI performs better than HI, meaning that the fuzzy control switcher initiative is at least as good as operator's judgment in switching LOA (i.e. compared to the HI of \cite{Chiou2016}) in the context of this experiment. The fact that MI outperforms RI possibly indicates learning effects in the primary task and the LOA switching. If that is the case, then it is due to the fact that we used the same participants in an identical experimental setup which traded the between-participants variation for the participant expertise as a confounding factor.

In terms of secondary task performance, MI and RI outperform HI. The secondary task completion time for MI and RI is faster than HI and on the same level of performance as the baseline condition (i.e. secondary task conducted in isolation from the primary task). This can be explained by two possible causes: a) secondary task learning effects; b) the fact that operators were feeling confident enough about the robot to neglect it and focus on the secondary task, given its capabilities to take initiative and alert the operator. The latter is reinforced by anecdotal evidence as an informal discussion with the participants revealed that most of them trusted the EMICS to take control and progress towards the navigation goal if needed. As one of the participants put it ``even if I completely neglect the robot, at least it will do something meaningful".

NASA-TLX showed that RI was perceived as marginally easier (i.e. less workload) than MI; and both MI and RI were perceived as significantly easier than HI. This can be an indication of the extra cognitive overhead that operators might have when they need to switch LOA based on judgment (e.g. HI). It is also an indication that having operators knowing that the EMICS will help them if needed, makes the task seem as easier. This is particularly true about the RI, as reflected in the NASA-TLX scores. Operators were restricted not to switch LOA themselves and thus, were not concerned about the LOA switching, just complied with EMICS's initiative. This is in contrast to MI in which operators still had to do some thinking about switching LOA.

There was no statistical difference in the average time spent in autonomy between HI, MI, and RI. Despite that fact, the time spent in autonomy for MI and RI was correlated with the primary task performance, contrary to HI. This possibly has to do with the fact that in MI and RI the time spent in each LOA (i.e. teleoperation and autonomy) was almost equally split. It is an indication that the operator and the EMICS deemed that both LOAs were required in a similar proportion to achieve good performance in the experiment.

Regarding the number of LOA switches in MI and HI, not only they were on the same level but they were also highly and positively correlated. The positive correlation means that participants followed similar patterns in the \textit{number of LOA switches} both in HI and in MI. For example, participants that switched LOA very frequently in HI, also switched LOA very frequently in MI and vise versa for low frequency. This reinforces the findings of \cite{Chiou2016_AAAI} that operators not only switch LOA based on reasons beyond performance, but also that personality traits play potentially a big role. Furthermore, the idea that some of the operator's LOA switching is redundant, is further reinforced by the fact LOA switches in RI were much less frequent than HI and MI. 

Lastly, out of the $13.6$ LOA switches in MI, $1.3$ were initiated by the EMICS. This means that despite operators' proven LOA switching capabilities and learning effects, there are circumstances in which the EMICS is successfully contributing by taking initiative. However, for some participants (4 out of the 16), the EMICS did not contribute any LOA switches. This possibly means that the fuzzy MI control switcher (i.e. the EMICS) is successful at choosing not to switch LOA when a switch is not needed. Given the variety of operator styles observed (i.e. high frequency and low frequency switches), this is not trivial, and is potential evidence that the control switcher is able to cope with different driving styles using the same parameter set.

\subsection{Summary of main findings}
\label{section:summary_second_exp}
This experiment offered several contributions and important insights into MI control. First, our Expert-guided MI Control Switcher (EMICS), both in MI and in RI conditions positively contributed towards performance and towards overcoming the degrading factors. This is especially true in the case of the RI condition in which the operator was not allowed to switch LOA. Hence, any learning effects regarding timely LOA switching did not affect the results. The small number of LOA switches in RI, and the small standard deviation compared to the rest of the conditions, is evidence that the control switcher's performance in LOA switching was consistent throughout the RI condition.

Second, the experiment provided evidence that when the EMICS is able to take initiative (i.e. MI and RI conditions), operators experience reduced workload. This is supported by NASA-TLX findings and also by the anecdotal evidence as discussed in Section \ref{section:learning_discussion}. Improved performance in the secondary task compared to HI can be partially explained by the lower workload, but without excluding the potential contribution of learning effects.

A human-factors explanation for the improved performance in MI and RI, both in primary and secondary tasks, as well as for reduced workload, may come from the operator's attentional model developed by Johnson et al. \cite{Johnson2017}. They model attention, workload, and performance across automation mode transitions in aviation and aerospace systems. In the HI condition operators have the extra cognitive overhead of having to switch LOA based on their judgment. MI and RI conditions reduce this overhead. As a result, the operator has more cognitive resources free which in turn can be re-allocated more efficiently via attention in the task at hand after a LOA switch. This improves performance as predicted by the model of Johnson et al. \cite{Johnson2017}. 

Lastly, the positive correlation in the number of LOA switches between MI and HI is an important finding suggesting that operators follow the same style in controlling the robot regardless of the fact that the EMICS can take initiative. This means that the operators have not changed their strategy in order to cope with the new system.

\section{Experiment 2: evaluation using real robot and test environment}
\label{experiment3}

The experiment presented in this section was designed to investigate how well the EMICS could generalize. Generalization in this context means that the EMICS should be able to function well (i.e. deliver similar performance to previous evaluation) in a different setting and under different conditions than the ones that led to its design (e.g. in the previous simulated environment). This is achieved by conducting an experiment with a real robot in a less controlled setting. More specifically the experiment aims to: a) provide evidence that the EMICS can generalize; b) factor out any learning effects that might have affected the results of the previous experiment; c) shift our experimental paradigm towards more complex, realistic environments and tasks.

\subsection{Experimental setup - apparatus, robot test arena, and control modes}

\begin{figure*}
		\centering
		\begin{subfigure}[b]{0.4\textwidth}
			\centering
			\includegraphics[width=\textwidth]{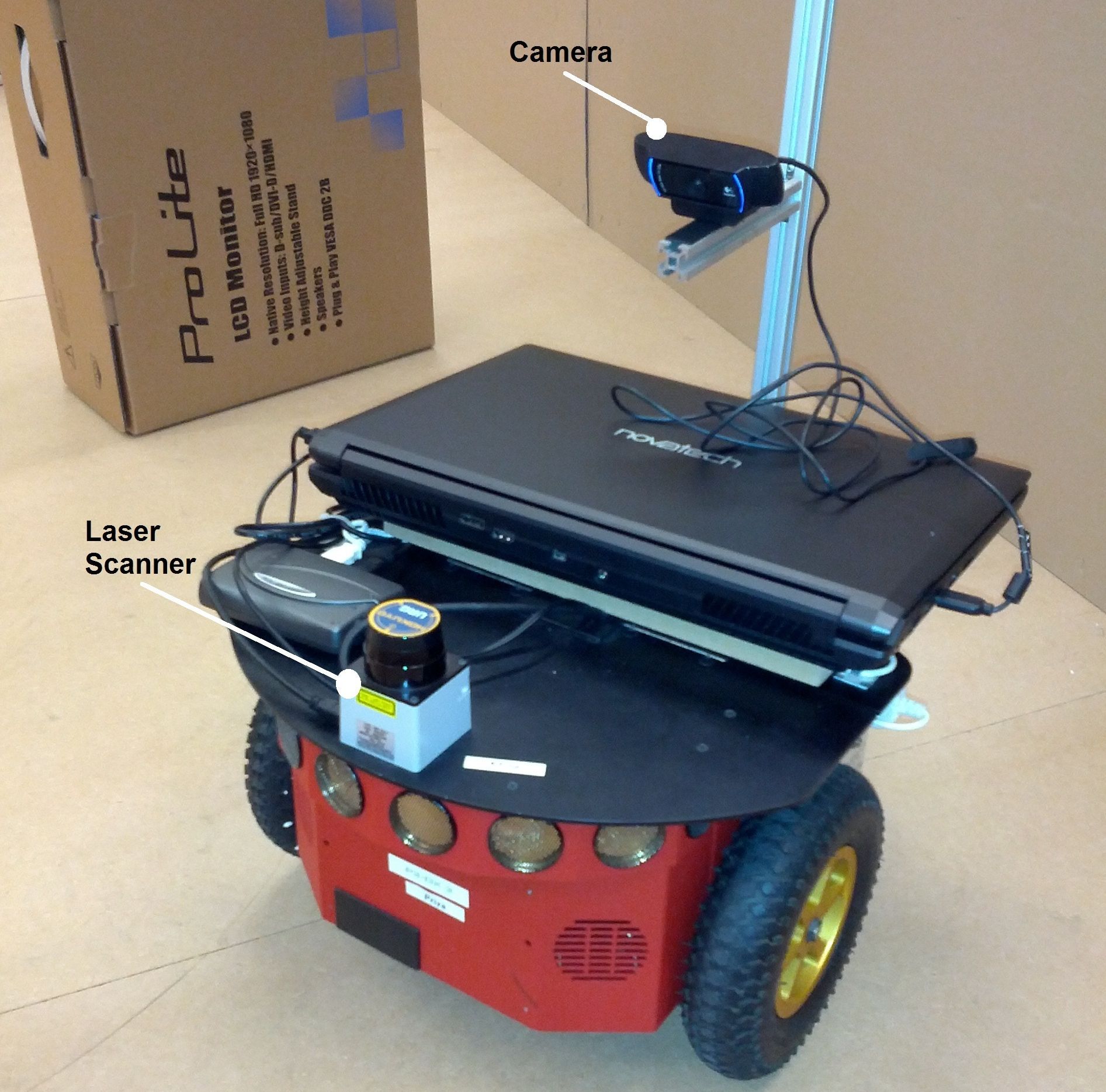}
			\caption{}
			\label{subfig:pioneer_exp3}
		\end{subfigure}
		\hfill
		\begin{subfigure}[b]{0.59\textwidth}
			\centering
			\includegraphics[width=\textwidth]{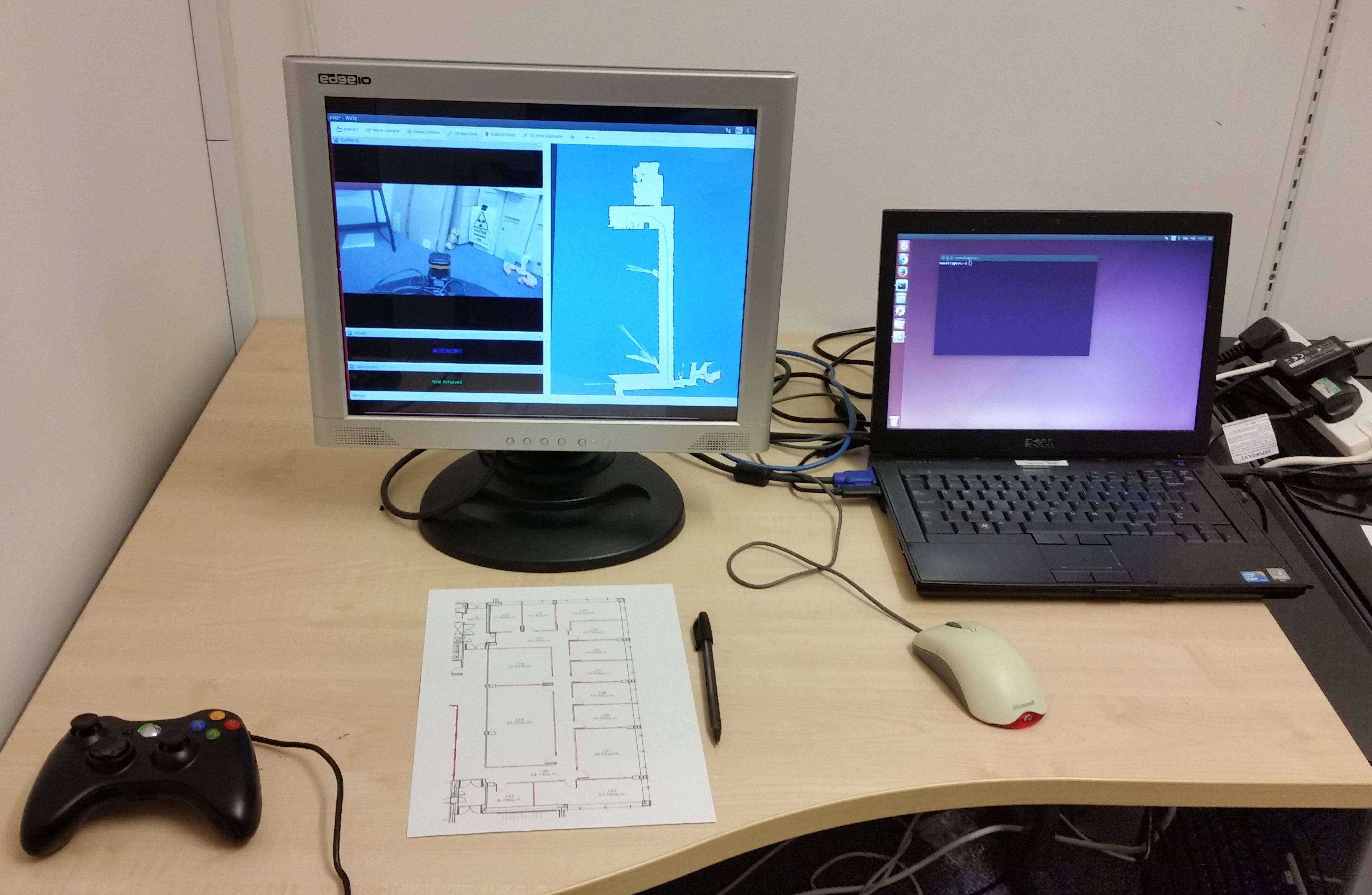}
			\caption{}
			\label{subfig:OCU_exp3}
		\end{subfigure}
		\hfill
		\caption{ \ref{subfig:pioneer_exp3}: The pioneer 3DX robot used in the experiment. \ref{subfig:OCU_exp3}: the Operator Control Unit (OCU), composed of a laptop, a joypad, a mouse and a screen showing the control interface. The same OCU was used in all variable autonomy experiments. Note the floor plan in front of the screen, used for the secondary task.}
		\label{fig:robot_ocu_exp3}
	\end{figure*}

\begin{figure*}
		\centering
		\begin{subfigure}[b]{0.69\textwidth}
			\centering
			\includegraphics[width=\textwidth]{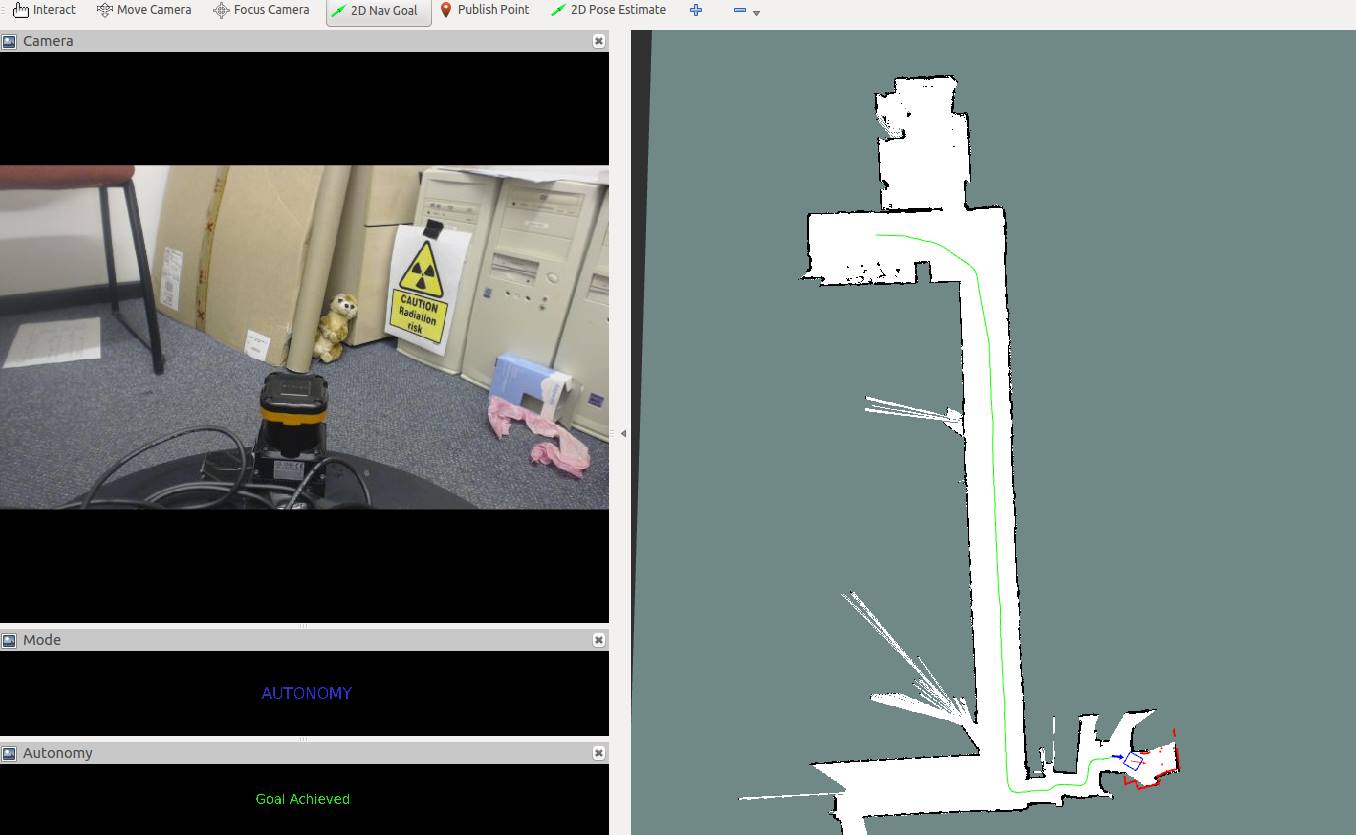}
			\caption{}
			\label{subfig:interface_exp3}
		\end{subfigure}
		\hfill
		\begin{subfigure}[b]{0.3\textwidth}
			\centering
			\includegraphics[width=\textwidth]{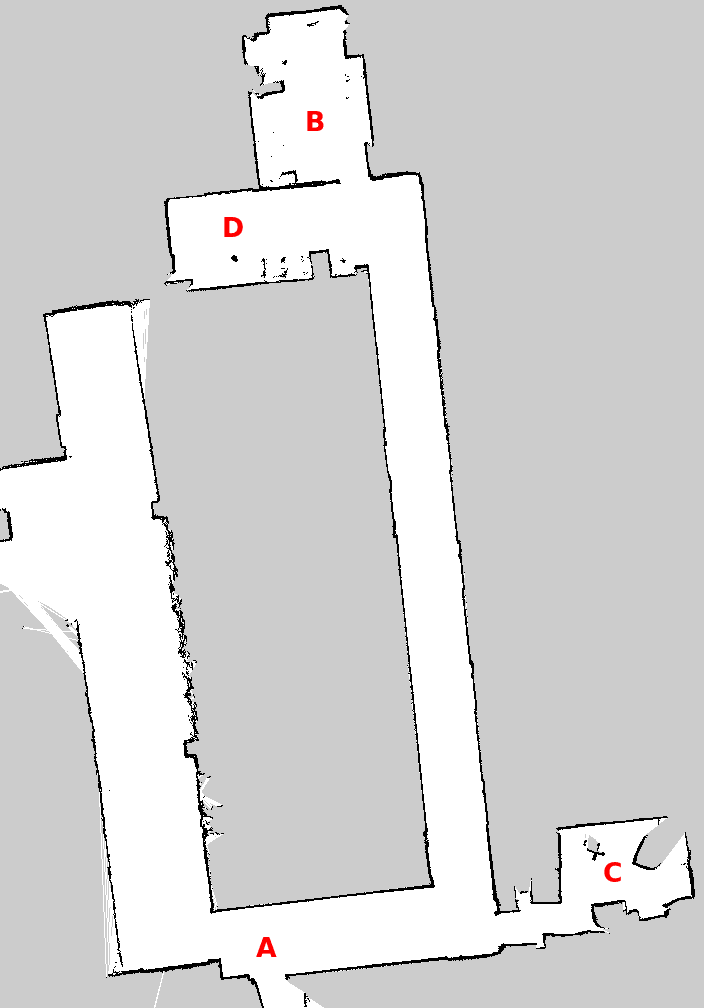}
			\caption{}
			\label{subfig:map_exp3}
		\end{subfigure}
		\hfill
		\caption{ \textbf{\ref{subfig:interface_exp3}:} The control interface as presented to the operator. \textbf{Left}: video feed from the camera, the control mode in use and the status of the navigation goal. \textbf{Right}: The map showing the position of the robot (blue footprint and red arrow), the current goal (blue arrow), the AI planner path (green line), the obstacles' laser reflections (red) and the walls (black). \textbf{\ref{subfig:map_exp3}:} The SLAM map of the arena as displayed to the human operator on the interface. Operators had to navigate in turn from point A; to B; to C; to D; and then back again to point A.}
		\label{fig:map_floor_exp3}
	\end{figure*}

In the experiment described here, we used identical software; variable autonomy control switchers; interface (see Figure \ref{subfig:interface_exp3}); and OCU (see Figure \ref{subfig:OCU_exp3}) to our previous experiment in simulation (see Section \ref{section:experiment2_2}). The robot used was a Pioneer-3DX equipped with a laser range finder and a camera (see Figure \ref{subfig:pioneer_exp3}). Operators controlled the robot remotely, from a separate location, via the control interface. Any situation-awareness (SA) was solely gained from the control interface. The communications link between the robot and the OCU was achieved via WiFi.

Our system offers the same two LOAs as in our previous experiment: \textbf{teleoperation} and \textbf{autonomy}. Three different control modes were tested in the experiment described here: 1) pure \textbf{teleoperation}, in which the operator was restricted to using only teleoperation LOA; 2) \textbf{Human-Initiative (HI)}, in which the operator could dynamically switch between the teleoperation and autonomy LOA using a button press; 3) \textbf{Mixed-Initiative (MI)}, in which both the operator and the EMICS had the ability and authority to dynamically switch between autonomy and teleoperation LOA.

\begin{figure*}
		\centering
		\begin{subfigure}[b]{0.15\textwidth}
			\centering
			\includegraphics[width=\textwidth]{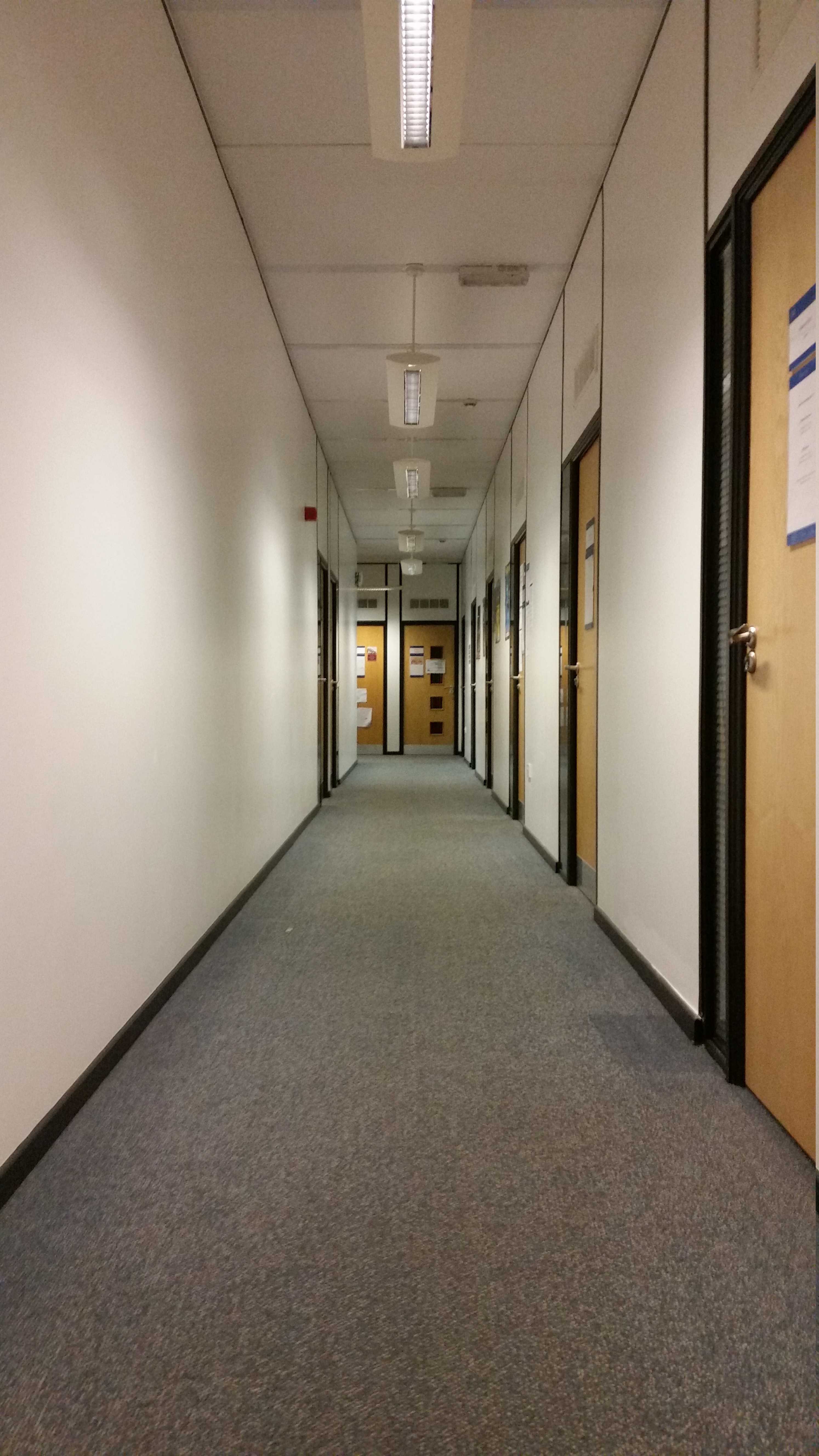}
			\caption{}
			\label{subfig:corridor_exp3}
		\end{subfigure}
		\hfill
		\begin{subfigure}[b]{0.3\textwidth}
			\centering
			\includegraphics[width=\textwidth]{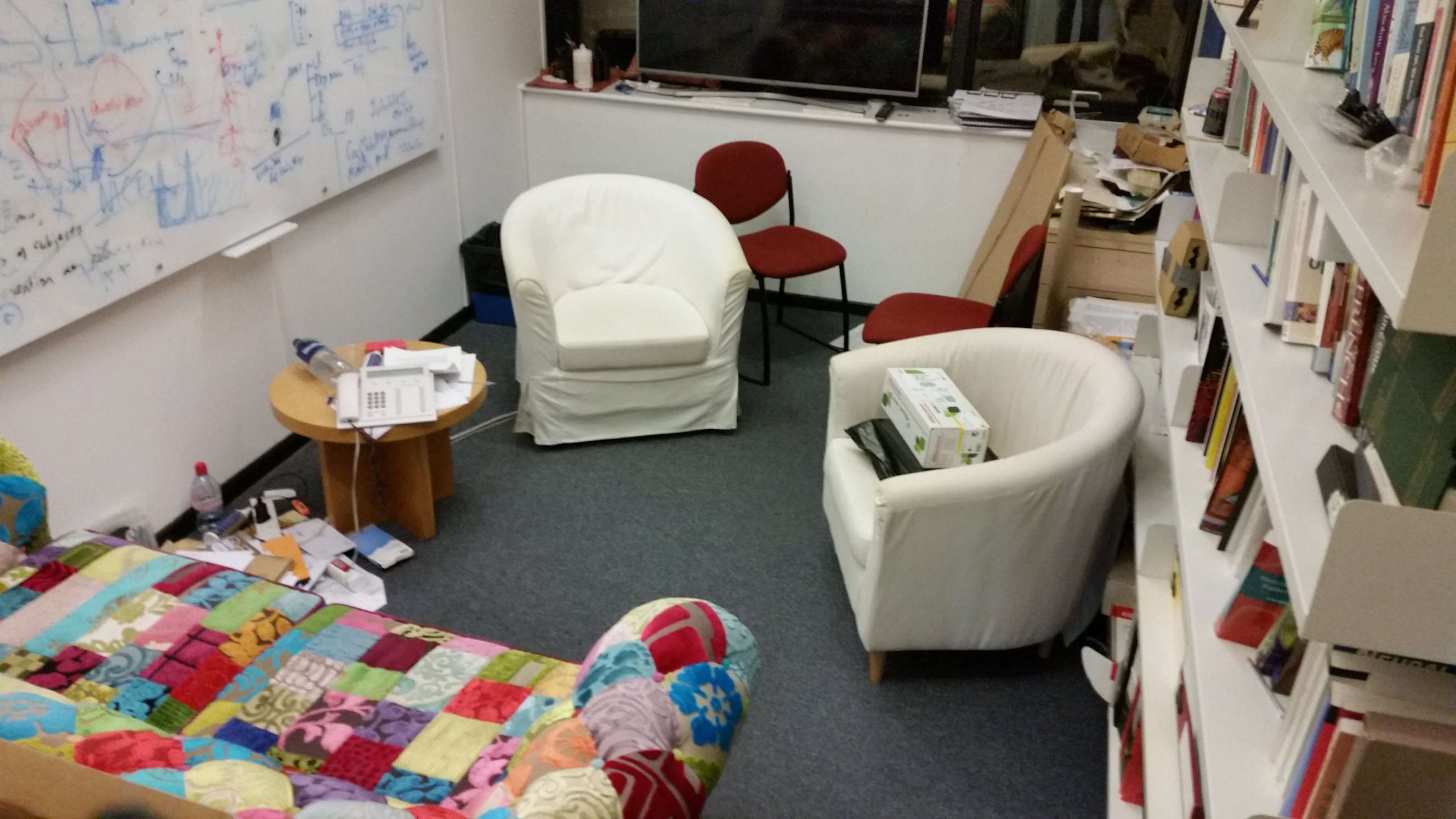}
			\caption{}
			\label{subfig:office_exp3}
		\end{subfigure}
		\hfill
		\caption{The first floor of School of Computer Science, University of Birmingham building was used as the arena for the USAR experiment. \textbf{\ref{subfig:corridor_exp3}:} The long corridor that connected the search areas (i.e. offices). \textbf{\ref{subfig:office_exp3}:} One of the offices used as a search area.}
		\label{fig:cs_arena_exp3}
	\end{figure*}

\begin{figure*}
		\centering
		\begin{subfigure}[b]{0.37\textwidth}
			\centering
			\includegraphics[width=\textwidth]{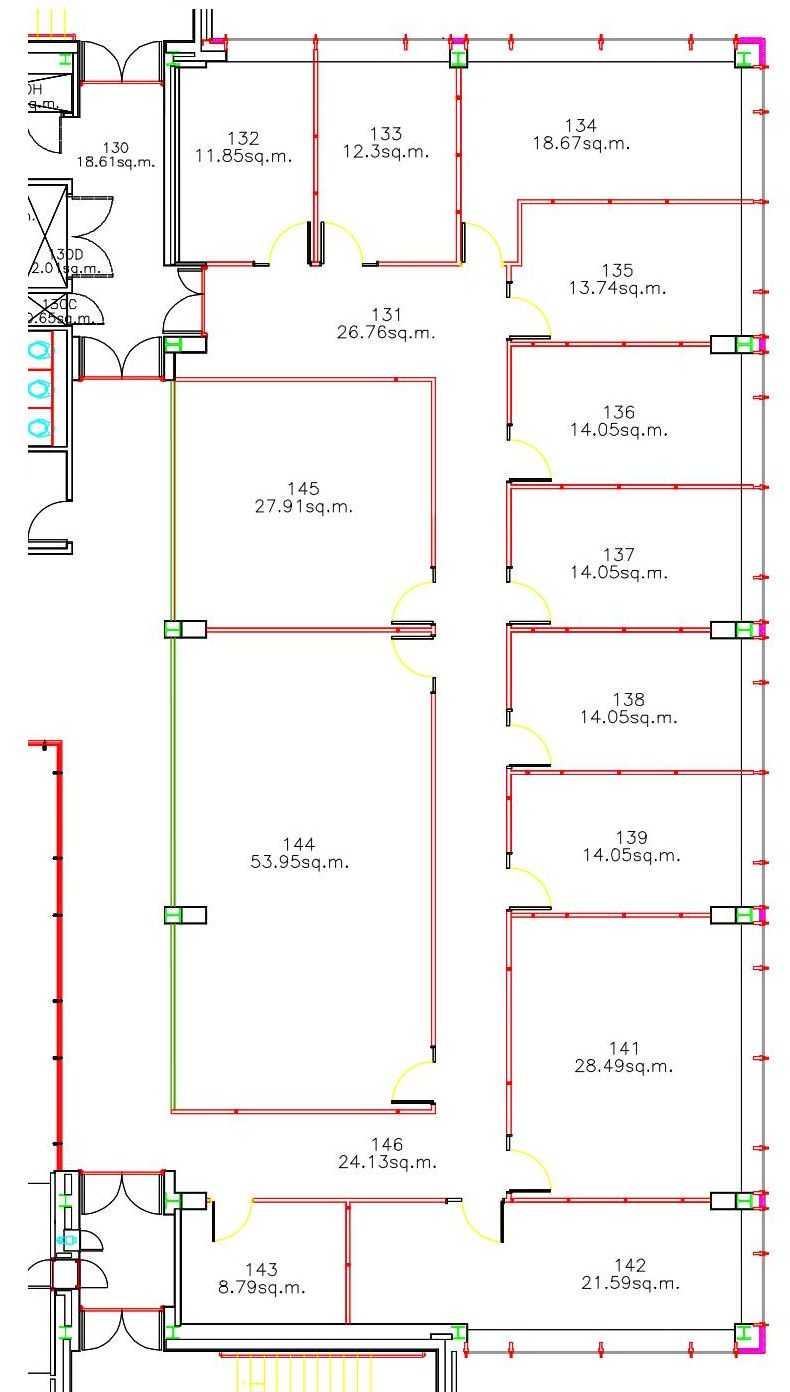}
			\caption{}
			\label{subfig:floorplan}
		\end{subfigure}
		\hfill
		\begin{subfigure}[b]{0.46\textwidth}
			\centering
			\includegraphics[width=\textwidth]{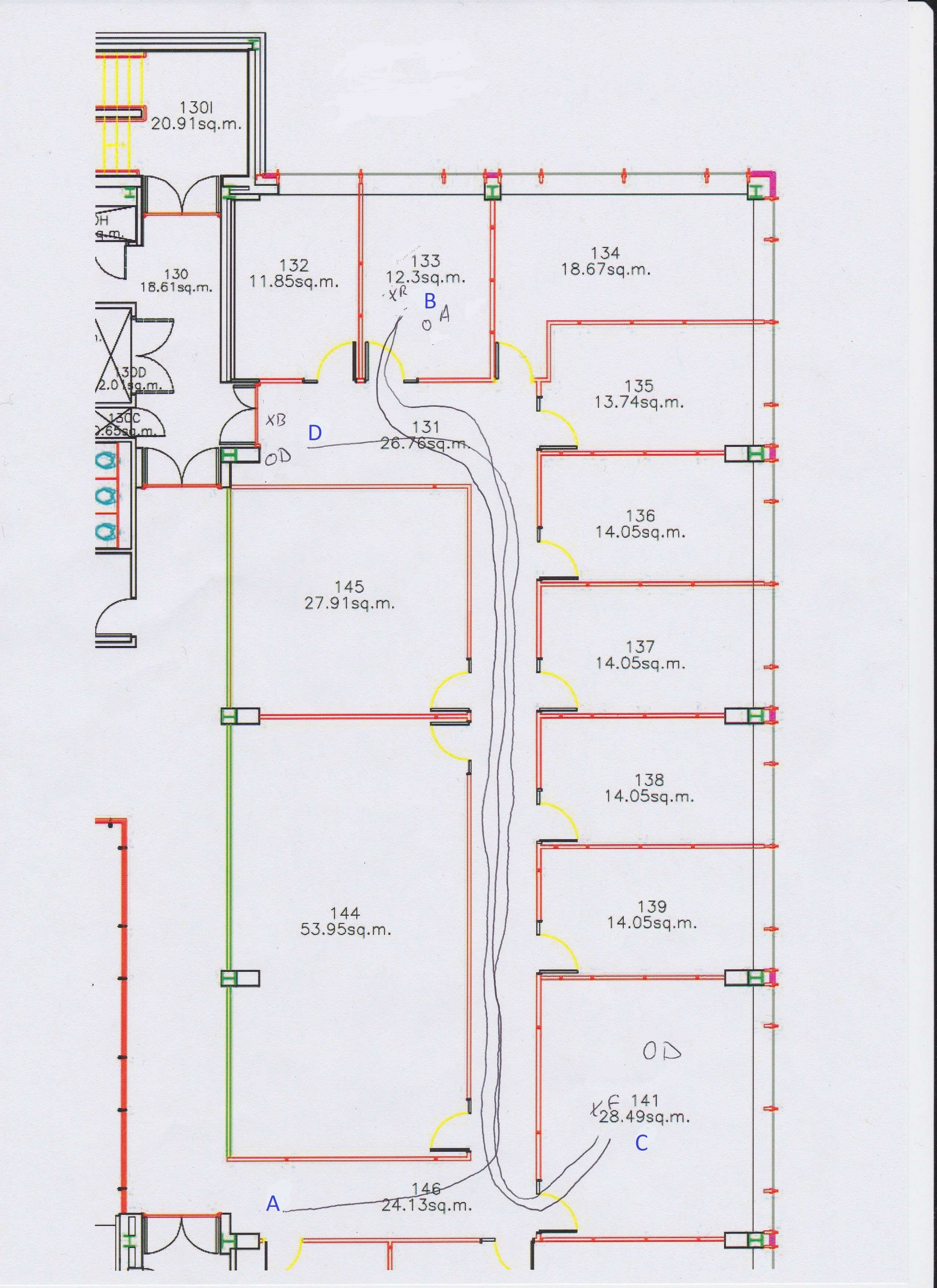}
			\caption{}
			\label{subfig:floorplan_annotated}
		\end{subfigure}
		\hfill
		\caption{ \textbf{\ref{subfig:floorplan}:} the floor plan, as kept in the university records, of the area that the experiment took place. This floor plan was printed and given to participants for the secondary task. \textbf{\ref{subfig:floorplan_annotated}:} the floor plan of the experiment area annotated by an operator during the secondary task. The handwritten annotations of the operator are not to be confused with the blue letters overlaid in order to denote the search areas for clarity in the figure.}
		\label{fig:floorplan_exp3}
	\end{figure*}

Part of the first floor of School of Computer Science, University of Birmingham building, was used as the arena for this experiment (see Figure \ref{subfig:map_exp3} and Figure \ref{subfig:floorplan}). More specifically, a long corridor (see Figure \ref{subfig:corridor_exp3}), 2 offices (see Figure \ref{subfig:office_exp3}) and an open space of approximately $101$ square meters in total were used. The experiment took place on weekends and out of hours in order to prevent human activity in the building being a confounding factor. 

\subsection{Tasks and performance degradation factors}
The overall theme of the experiment was an Urban Search and Rescue (USAR) scenario. In this scenario the robot operator had to remotely control the robot in the search zone (i.e. the building) and identify the positions and the statuses of victims and potential hazards. As is often the case in real operations, we assume that some prior knowledge of the building is given to the robot operator from the authorities or any relevant organization. This is represented in the experiment by the SLAM map that appears on the interface (see Figure \ref{subfig:map_exp3}) and by the floor plan used in the secondary task, as kept in the university's records (see Figure \ref{subfig:floorplan}). 

\begin{figure*}
		\centering
		\begin{subfigure}[b]{0.3\textwidth}
			\centering
			\includegraphics[width=\textwidth]{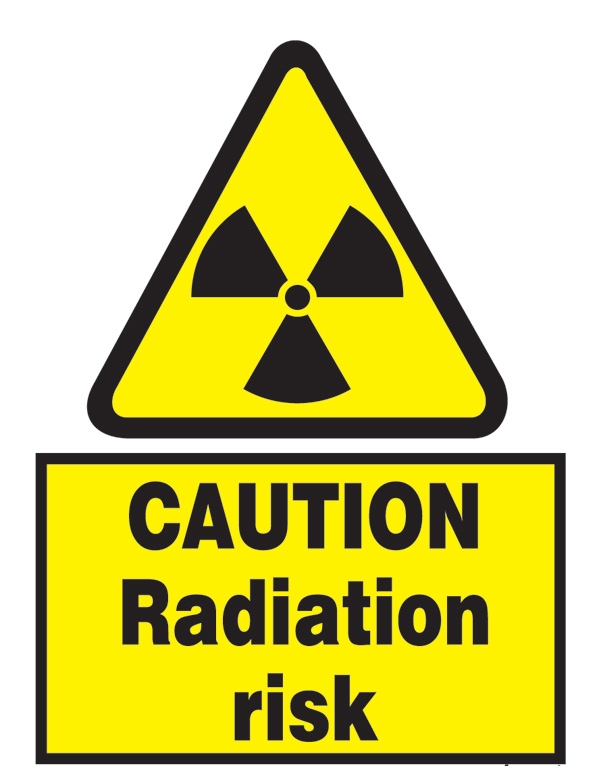}
		\end{subfigure}
		\hfill
		\begin{subfigure}[b]{0.3\textwidth}
			\centering
			\includegraphics[width=\textwidth]{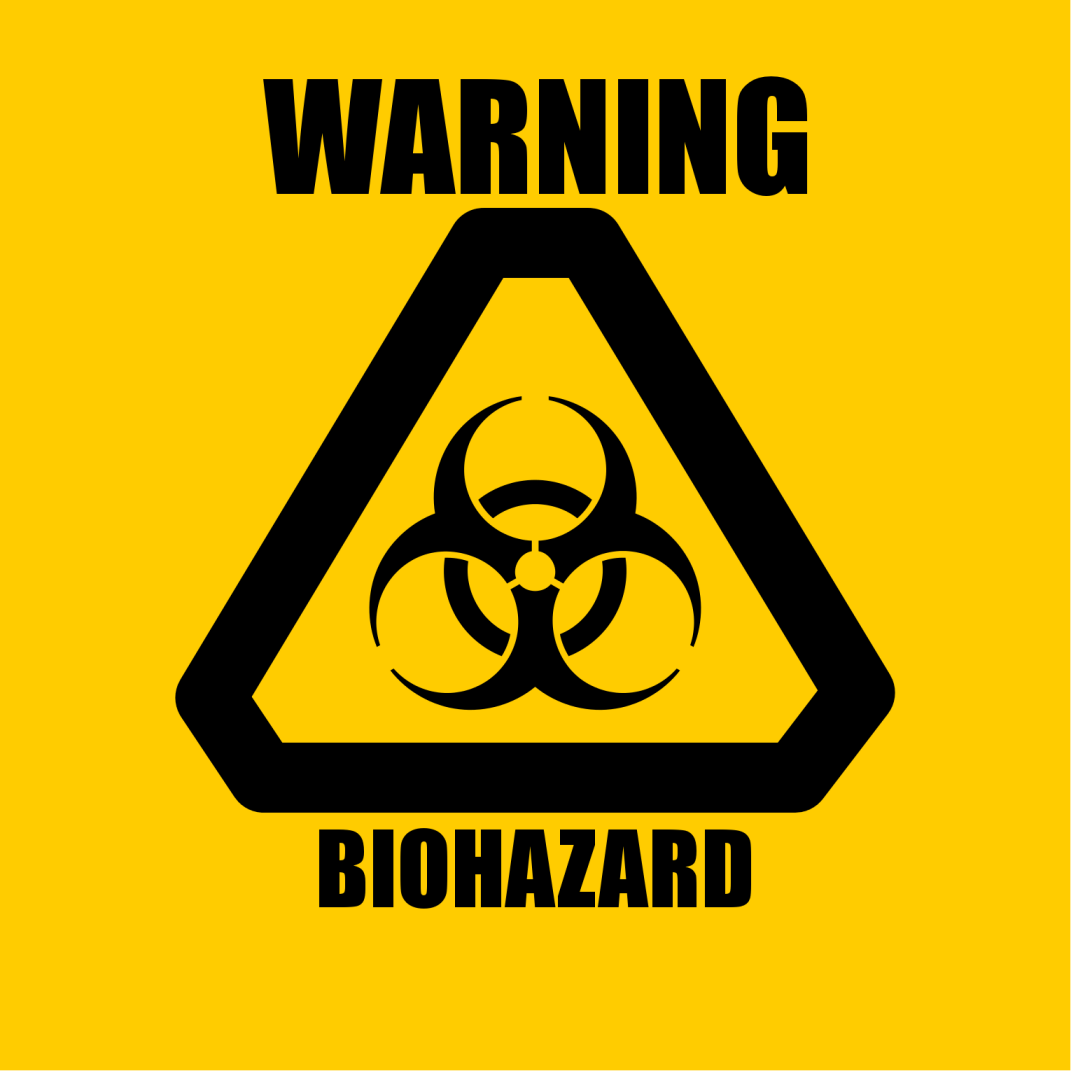}
		\end{subfigure}
		\hfill
		\caption{Two of the hazard signs used in order to denote bio-hazard and radiation risk.}
		\label{fig:hazards_exp3}
	\end{figure*}

\begin{figure*}
		\centering
		\begin{subfigure}[b]{0.35\textwidth}
			\centering
			\includegraphics[width=\textwidth]{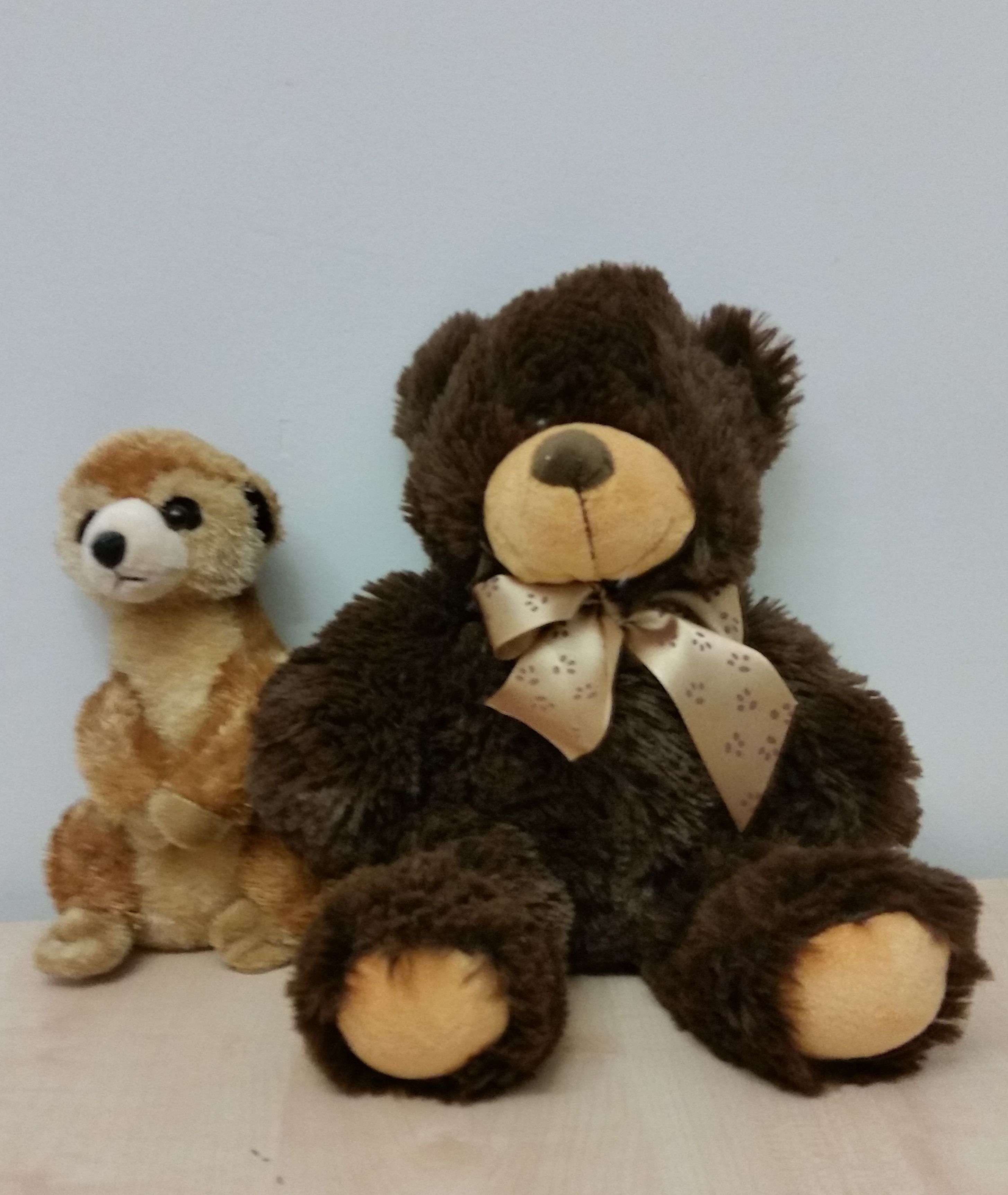}
			\caption{}
			\label{subfig:victims}
		\end{subfigure}
		\hfill
		\begin{subfigure}[b]{0.4\textwidth}
			\centering
			\includegraphics[width=\textwidth]{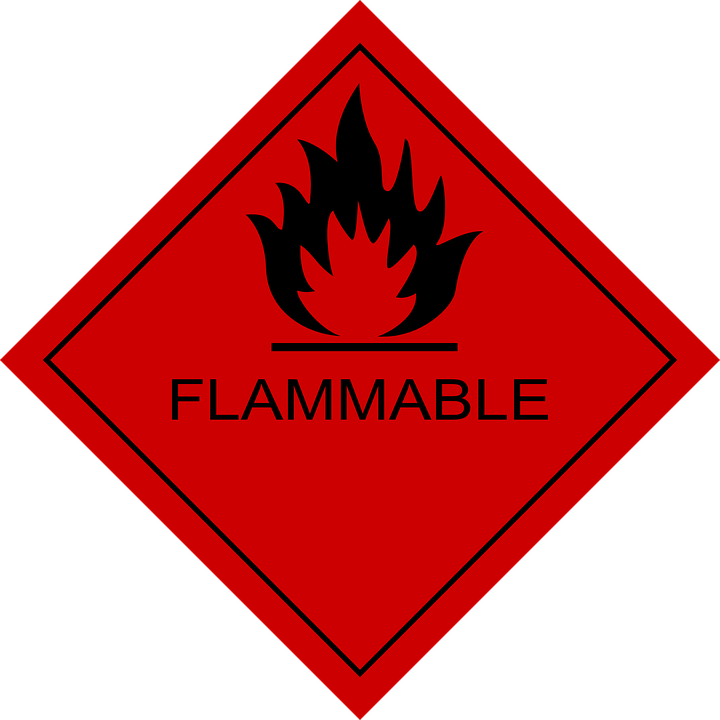}
			\caption{}
			\label{subfig:flammable}
		\end{subfigure}
		\hfill
		\caption{ \textbf{\ref{subfig:victims}:} The stuffed animals representing the victims of the USAR scenario. A meerkat was representing a victim which is alive and a teddy-bear a victim which is dead. \textbf{\ref{subfig:flammable}:} The hazard sign used to denote flammable materials risk.}
		\label{fig:flammable_victims}
	\end{figure*}

The primary task was to navigate the robot between the different areas in the map in a predefined order. From point A to point B, C, D and then back to point A (see Figure \ref{subfig:map_exp3}). In every one of those areas/points (excluding A) one victim and one hazard sign was placed. The victims were represented by stuffed animals. A meerkat represented an alive victim and a teddy-bear a dead victim (see Figure \ref{subfig:victims}). Hazards were represented by 3 commonly used hazard signs for flammable materials, radiation, and bio-hazard (see Fig \ref{fig:hazards_exp3} and Figure \ref{subfig:flammable})). Inside the search areas the robot had to stop in the center of the room. Then the operator would have to identify and memorize the position and status of the victim and the sign. Both the victim and the sign were visible from the center of the area and in $360$ degrees around the robot (i.e. the victim and the sign could be anywhere around the central position, but visible); no further exploration was needed.

In addition to the primary task, the operator had to perform a secondary task every time the robot exited one of the search areas. This secondary task was designed to induce additional workload in the operator and degrade performance on the primary task. A pen and a paper displaying the floor plan was placed in-front of the operator (see Figure \ref{subfig:OCU_exp3}). When the operator was asked to perform the secondary task by the experimenter, they had to sketch on the floor plan paper (see Figure \ref{subfig:floorplan_annotated}): a) the position of the victim denoted by a small x; b) the status of the victim denoted by a letter (A for alive, D for dead); c) the position of the hazard denoted by a small o; d) the status of the hazard denoted by a letter (R - radiation, F - flammable, and B - bio-hazard); e) the path that the robot followed from the previous visited point/area. The pen and the floor plan were placed in front of the operator and in a central position (see Figure \ref{subfig:OCU_exp3}). This way individual differences regarding handedness (i.e. left vs right handed participants) did not bias the results. Similar tasks to the one described here are typically required from robot operators in real world disaster response \cite{Murphy2004} as they are asked to sketch similar information for the SAR team. The speed and ability of the operators to annotate hazards, victims and paths, along with their correct status and respective positions, were the measures of secondary task performance.

The robot performance was degraded by two different degradation factors. The first factor was a box placed in one of the offices. This box was not part of the prior knowledge (i.e. the box was not in the SLAM map). The box also narrowed the entrance to the office. For these two reasons navigation performance would degrade (i.e. the robot is either moving very slowly or is stuck), as the robot would try to plan a new path through this very tight passage and not through the obstacle. The second performance degradation factor was naturally occurring noise in the laser sensor in certain parts of the arena. This noise was due to shiny surfaces deflecting the laser's beams and it was not controlled by the experimenter. However, due to the high and semi-systematic frequency of its appearance in the specific area of the map, it was adopted as part of the experimental design. Both of these factors, naturally occurring laser noise and unknown obstacles, often occur in real scenarios as the environments are dynamic. This adds to the realism of the experiment.  

Lastly, in one of the offices the WiFi signal was weak. As a result, delays in the control commands and in SA updates in the interface (e.g. location in map, video feedback etc) occurred. This was systematic throughout all of the participants and trials. Hence, it did not constitute a confounding factor, while contributing further to the realism of the experiment.

\subsection{Participants and experimental design}
A total of 12 volunteers participated with the majority been males in their early 20s as they were recruited from the School of Computer Science’s population. A within-groups experimental design was followed as every participant performed one trial for each of the 3 control modes (i.e. teleoperation, MI, and HI). The order of the three trials was rotated between all the different permutations for different participants. This counterbalancing technique was used in order to prevent learning and fatigue effects from introducing confounding factors to the results, since every participant performed all three trials. Additionally, for the secondary task, the signs, the victims, and their positions were randomized in every trial. Again, this was in order to eliminate any learning effects. A prior experience questionnaire showed, as in our previous experiments, that the majority of the participants were experienced in playing video games. Moreover, 8 out of the 12 participants participated in our previous experiment. 

Participants underwent extensive standardized training, similar to our previous experiment. Due to space constraints a sub-region of the experiment's area was used for the training. In order for the participants to proceed with the experiment they had first to demonstrate their abilities by completing three standardized test trials, one for every control mode. These trials mimicked the actual experimental trials (i.e. same primary and secondary tasks). The training and the test trials ensured that all participants had attained a common minimum skill level in operating the robot. 

Participants were instructed to perform the primary task (controlling the robot to search the areas) as quickly and safely (i.e. avoiding collisions) as possible. Additionally they were told that when instructed to perform the secondary task (i.e. annotating information regarding victims, hazards, and paths), they should do it as quickly and as accurately as possible. They were explicitly told that they should give priority to the secondary task over the primary task and should only perform the primary task if the workload allowed. Lastly, participants were told that the best performing individuals in each trial would be rewarded with an extra gift voucher. However, they were not informed about how the best performing participants would be decided, as it would have had them biased towards specific factors. The purpose of this extra gift voucher was to provide an incentive for participants to achieve the best performance possible on both primary and secondary task.

At the end of each trial, participants had to complete an online NASA Task Load Index (NASA-TLX) questionnaire. As in our previous experiments, NASA-TLX was used to rate the level of difficulty and workload the participants experienced during each trial.

\subsection{Results}
A repeated measures one-way ANOVA was used. The independent variable was the control mode with three levels: teleoperation; HI; and MI. In the cases that sphericity assumption was violated a Greenhouse-Geisser correction was used with the ANOVA. For pairwise comparisons after a significant ANOVA result, Fisher's least significant difference (LSD) test was used to determine the conditions that differed. Similar to the rest of the paper, we consider a result to be significant when it yields a $p$ value less than $0.05$. We also report on the statistical power of the results and on the effect size using $\eta^2$. The detailed statistical calculations are reported in Table \ref{table:results_exp_3}.

\begin{table*}[t]
\caption{Table showing ANOVA results and descriptive statistics for all the metrics. For the number of LOA switches, the t-test result is reported.}
	\centering
	\begin{tabular}{lll}
		\hline
		\textbf{metric}                                                           & \textbf{ANOVA control mode effect}                                                                               & \textbf{descriptive statistics}                                                                                                                               \\ \hline
		\begin{tabular}[c]{@{}l@{}}primary task\\ completion time\end{tabular}    & \begin{tabular}[c]{@{}l@{}}$F(2, 22) = 7.382$, $p < .01$, \\ $power > .9$, $\eta^2 = .402$\end{tabular}          & \begin{tabular}[c]{@{}l@{}}HI: $M = 327.9$ $sec$, $SD = 23.38$ \\ MI: $M = 349.5$ $sec$, $SD = 35.95$ \\ teleop: $M = 366.3$ $sec$, $SD = 36.26$\end{tabular} \\ \hline
		collisions                                                                & \begin{tabular}[c]{@{}l@{}}$F(2, 22) = .786$, $p > .05$,\\ $power < .8$, $\eta^2 = .067$,\end{tabular}           & \begin{tabular}[c]{@{}l@{}}HI: $M = 0.08$, $SD = .289$\\ MI: $M = 0.08$, $SD = .289$ \\ teleop: $M = 0.25$, $SD = 4.52$\end{tabular}                          \\ \hline
		\begin{tabular}[c]{@{}l@{}}primary task \\ score\end{tabular}             & \begin{tabular}[c]{@{}l@{}}$F(2, 22) = 8.724$, $p < .01$, \\ $power > .9$, $\eta^2 = .442$\end{tabular}          & \begin{tabular}[c]{@{}l@{}}HI: $M = 328.8$, $SD = 22.42$ \\ MI: $M = 350.3$, $SD = 34.9$ \\ teleop: $M = 368.8$, $SD = 35.57$\end{tabular}                    \\ \hline
		\begin{tabular}[c]{@{}l@{}}secondary task \\ completion time\end{tabular} & \begin{tabular}[c]{@{}l@{}}$F(2, 22) = 0.68$, $p > .05$, \\ $power > .85$, $\eta^2 = .058$\end{tabular}          & \begin{tabular}[c]{@{}l@{}}HI: $M = 54$ $sec$, $SD = 12.86$ \\ MI: $M = 54.8$ $sec$, $SD = 20.16$\\ teleop: $M = 60.3$ $sec$, $SD = 16.37$\end{tabular}       \\ \hline
		\begin{tabular}[c]{@{}l@{}}secondary task \\ errors\end{tabular}          & \begin{tabular}[c]{@{}l@{}}$F(2,22) = .789$, $p > .05$,\\ $power < .8$,  $\eta^2 = .067$\end{tabular}            & \begin{tabular}[c]{@{}l@{}}HI: $M = .83$, $SD = 1.03$ \\ MI: $M = 1.5$, $SD = 1.314$ \\ teleop: $M = 1.42$, $SD = 1.62$\end{tabular}                          \\ \hline
		\begin{tabular}[c]{@{}l@{}}NASA-TLX\\ scores\end{tabular}                 & \begin{tabular}[c]{@{}l@{}}$F(1.263,13.896) = 5.001$, $p < .05$, \\ $power = .603$, $\eta^2 = .313$\end{tabular} & \begin{tabular}[c]{@{}l@{}}HI: $M = 25.9$, $SD = 12.39$ \\ MI: $M = 29.9$, $SD = 14.94$ \\ teleop: $M = 38.9$, $SD = 17.4$\end{tabular}                       \\ \hline
		\begin{tabular}[c]{@{}l@{}}number of \\ LOA switches\end{tabular}         & $t(11) = -4.076 , p < 0.01$                                                                                      & \begin{tabular}[c]{@{}l@{}}HI: $M = 9.08$, $SD = 3.42$ \\ MI: $M = 15.08$, $SD = 5.73$\\controller in MI: $M = 3.42$, $SD = 1.83$\end{tabular}           \\ \hline
	\end{tabular}
	\label{table:results_exp_3}
\end{table*}

\begin{figure*}[t]
		\centering-
		\begin{subfigure}[b]{0.49\textwidth}
			\centering
			\includegraphics[width=\textwidth]{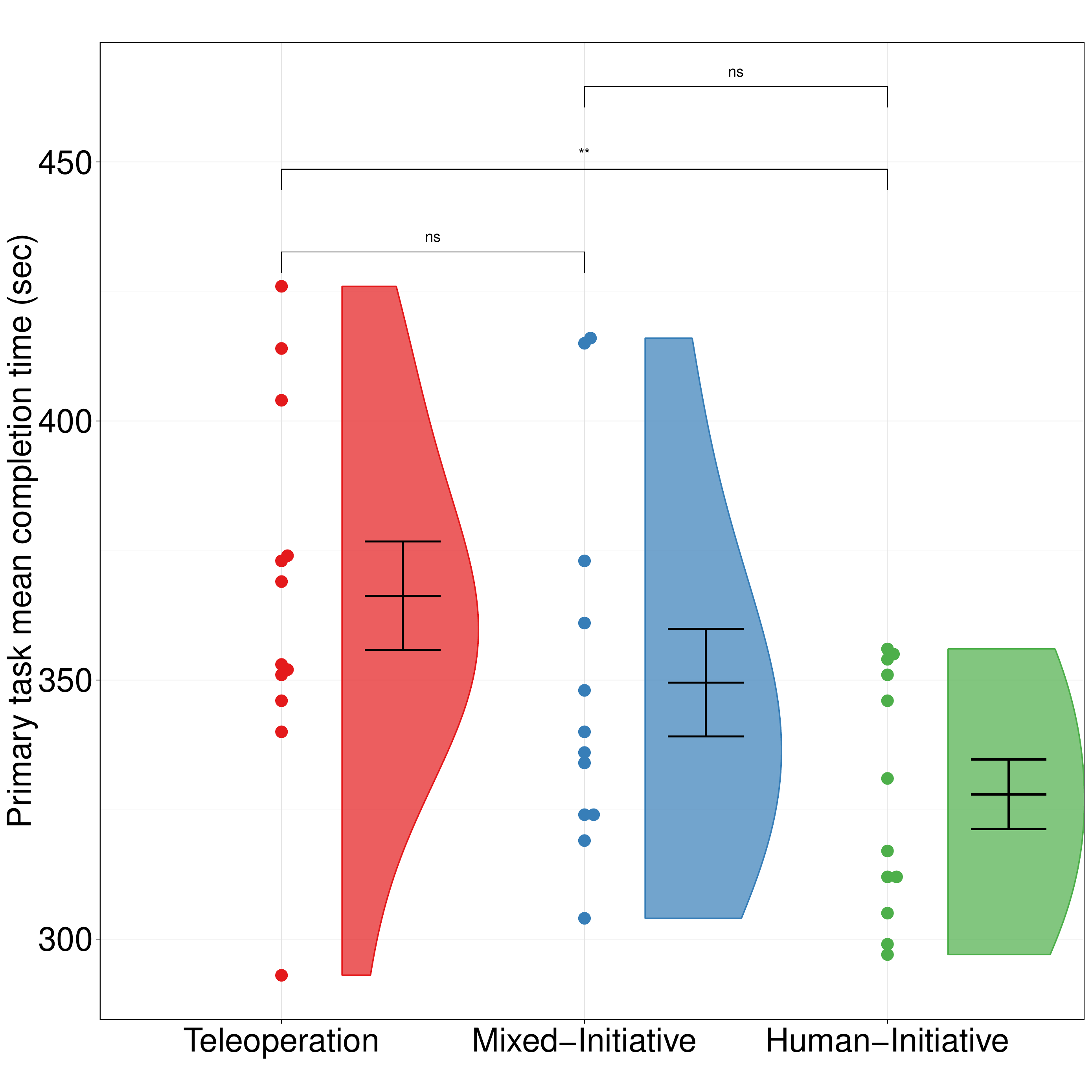}
			\caption{}
			\label{subfig:primary_time_exp3}
		\end{subfigure}
		\hfill
		\begin{subfigure}[b]{0.49\textwidth}
			\centering
			\includegraphics[width=\textwidth]{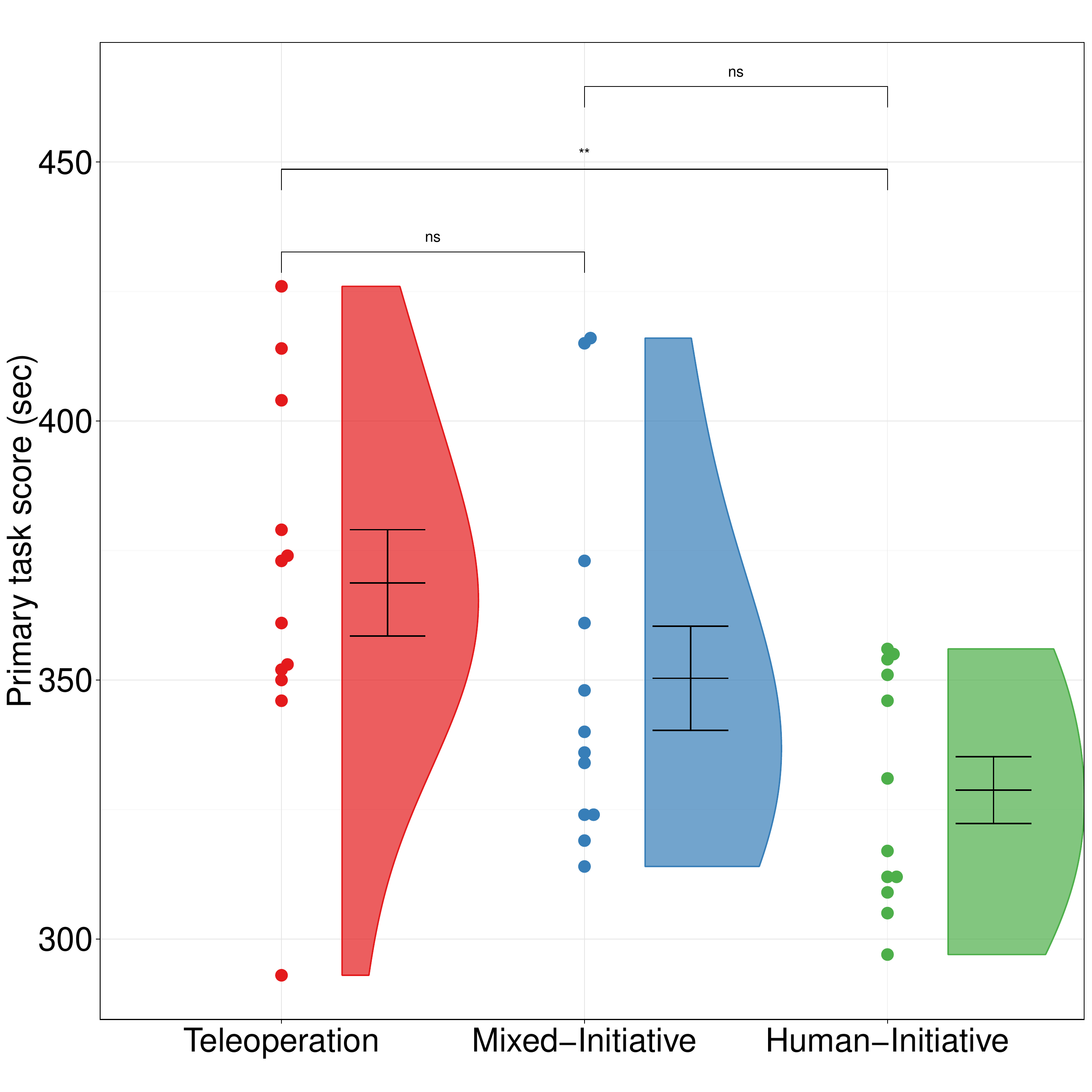}
			\caption{}
			\label{subfig:primary_score_exp3}
		\end{subfigure}
		\hfill
		\caption{\textbf{\ref{subfig:primary_time_exp3}:} Primary task mean time-to-completion. \textbf{\ref{subfig:primary_score_exp3}:} Primary task score combining time and collisions penalty. HI significantly outperformed teleoperation in primary task.}
		\label{fig:exp3_primary}
	\end{figure*}

ANOVA for \textit{primary task completion time} (see Figure \ref{subfig:primary_time_exp3}) showed overall significantly different means between HI variable-autonomy, MI and teleoperation. Pairwise comparison reveals that HI performed significantly better (i.e. lower mean completion time) than teleoperation (\textit{$p < .01$}). MI primary task completion time was statistically in the same level as teleoperation and HI (\textit{$p > .05$}). The effect of control mode on the number of \textit{collisions} was not significant. 

Similar to our previous experiments the \textit{primary task score} (see Figure \ref{subfig:primary_score_exp3}) was calculated in order to capture any speed-accuracy trade-offs in the primary task. The primary task score was calculated by adding a time penalty of \textit{$10$ $sec$} for every collision onto the primary task completion time for each participant. ANOVA analysis showed that control mode had a significant effect on the primary task score. As expected due to the small number of collisions, LSD pairwise tests showed very similar results with the primary task completion time. The HI control switcher significantly (\textit{$p < .01$}) outperformed pure teleoperation. The primary task score for MI control switcher was in the same level (i.e. no statistical difference found) as pure teleoperation and HI (\textit{$p > .05$}). 

\begin{figure*}
		\centering
		\begin{subfigure}[b]{0.49\textwidth}
			\centering
			\includegraphics[width=\textwidth]{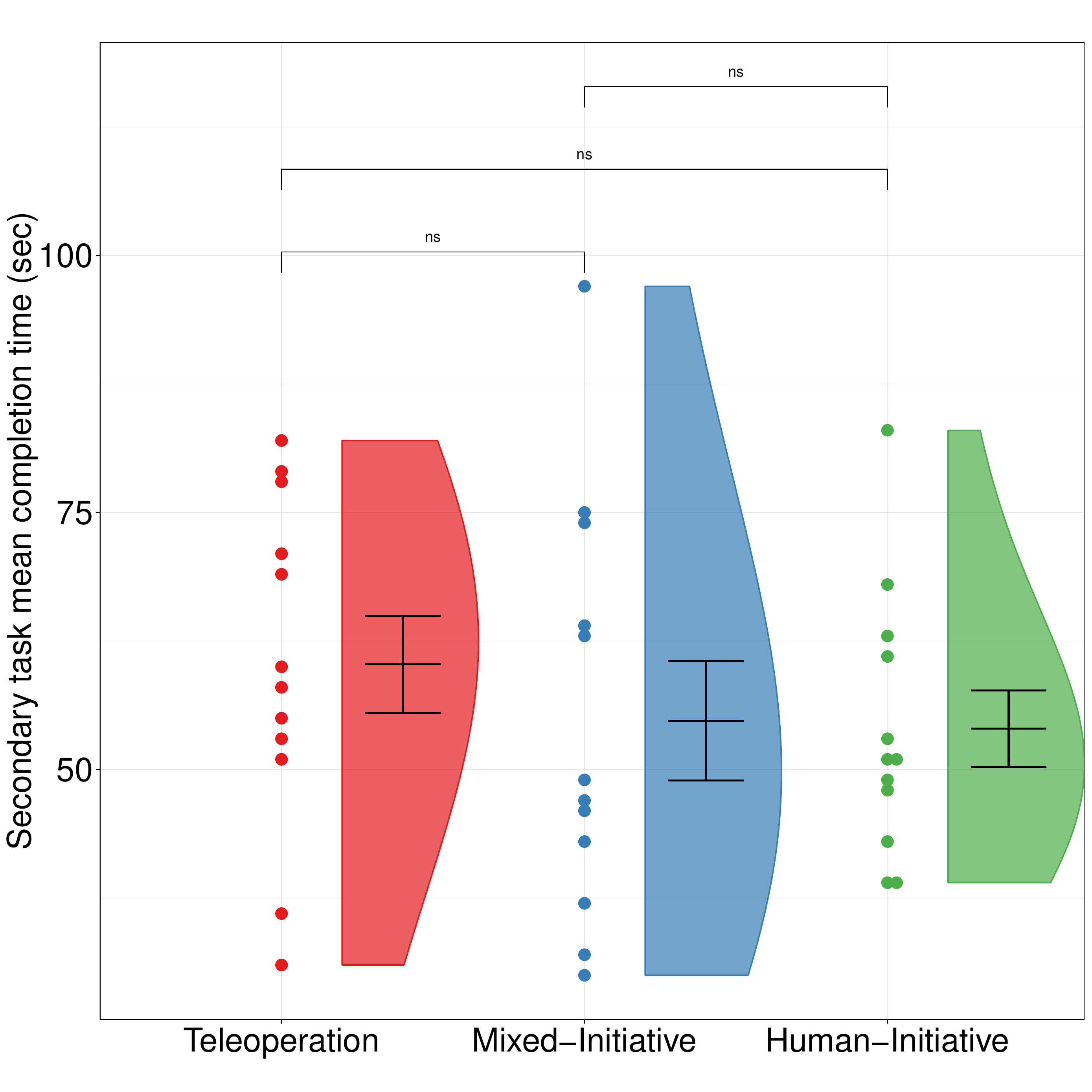}
			\caption{}
			\label{subfig:secondary_time_exp3}
		\end{subfigure}
		\hfill
		\begin{subfigure}[b]{0.49\textwidth}
			\centering
			\includegraphics[width=\textwidth]{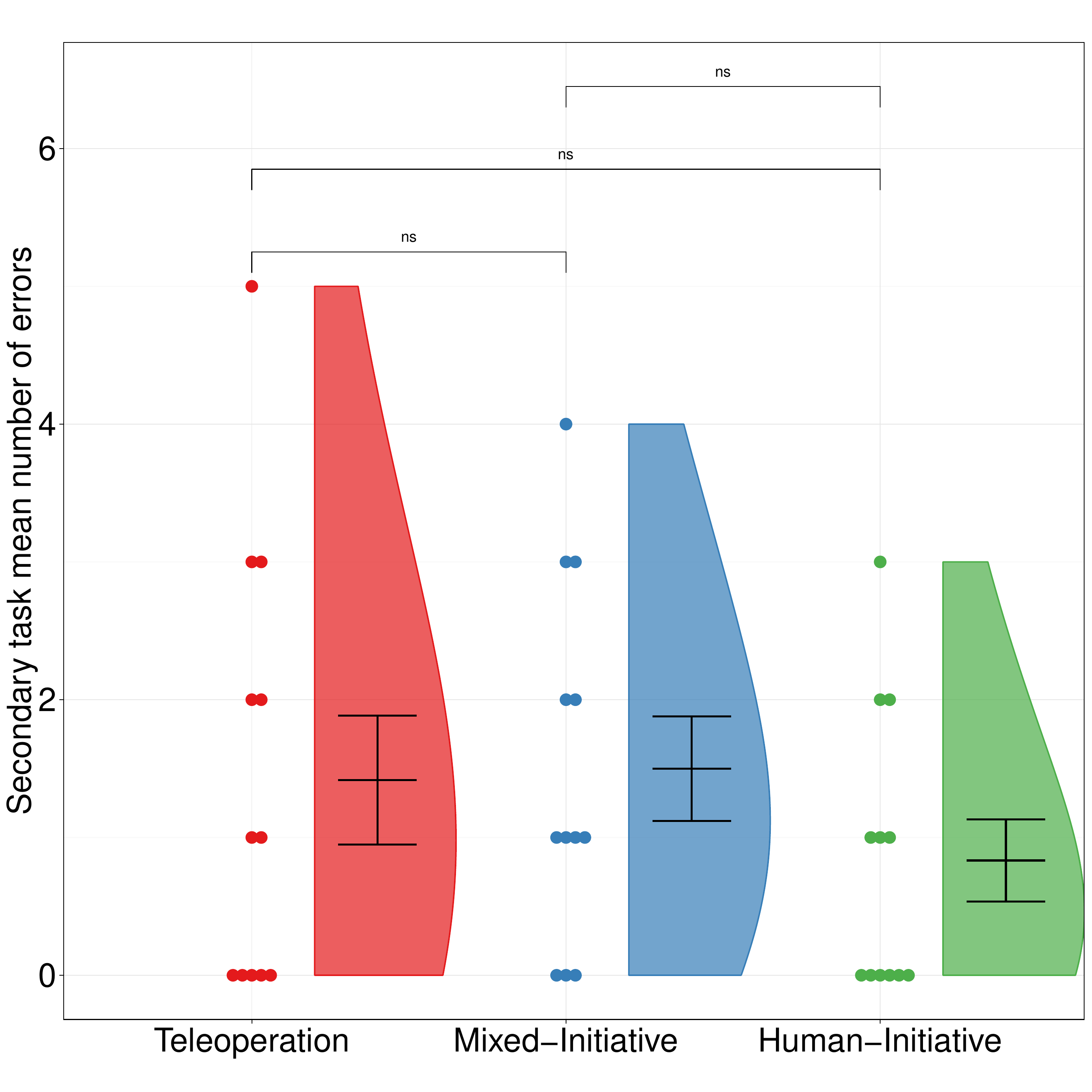}
			\caption{}
			\label{subfig:secondary_mistakes_exp3}
		\end{subfigure}
		\hfill
		\caption{\textbf{\ref{subfig:secondary_time_exp3}:} Secondary task completion time. \textbf{\ref{subfig:secondary_mistakes_exp3}:} Secondary task total number of errors for each trial.}
		\label{fig:secondary_task_exp3}
	\end{figure*}

\textit{Secondary task completion time} (see Figure \ref{subfig:secondary_time_exp3}) refers to the total time per trial that the participants took to complete the full annotation in the floor footprint. A full annotation is defined as a sketch that has annotated positions, statuses and paths for all the three search areas. ANOVA did not suggested a significant difference between the mean secondary task completion times for the different control modes.

The \textit{secondary task number of errors} was measured. A position error was defined an annotation not representing the true position of a victim or hazard with relative accuracy. A status error was defined an annotation not representing the correct status, e.g. victim annotated alive (i.e. meerkat) when is it in-fact dead (i.e. teddy-bear). A path error was defined: a) a path that was not annotated; b) a path that was heavily colliding with the walls in the floor plan; c) a not accurately depicted path. In a real situation the rescue team should be able to find the victims which are alive, be aware of any hazards and their nature. This should happen by following the path and the position annotations provided and by using common sense regarding space orientation. This is represented in our experiment by the secondary task errors. No significant differences were observed between the different control modes with respect to the number of secondary task errors (see Figure \ref{subfig:secondary_mistakes_exp3}) according to ANOVA.

\begin{figure*}[t]
		\centering
		\begin{subfigure}[b]{0.49\textwidth}
			\centering
			\includegraphics[width=\textwidth]{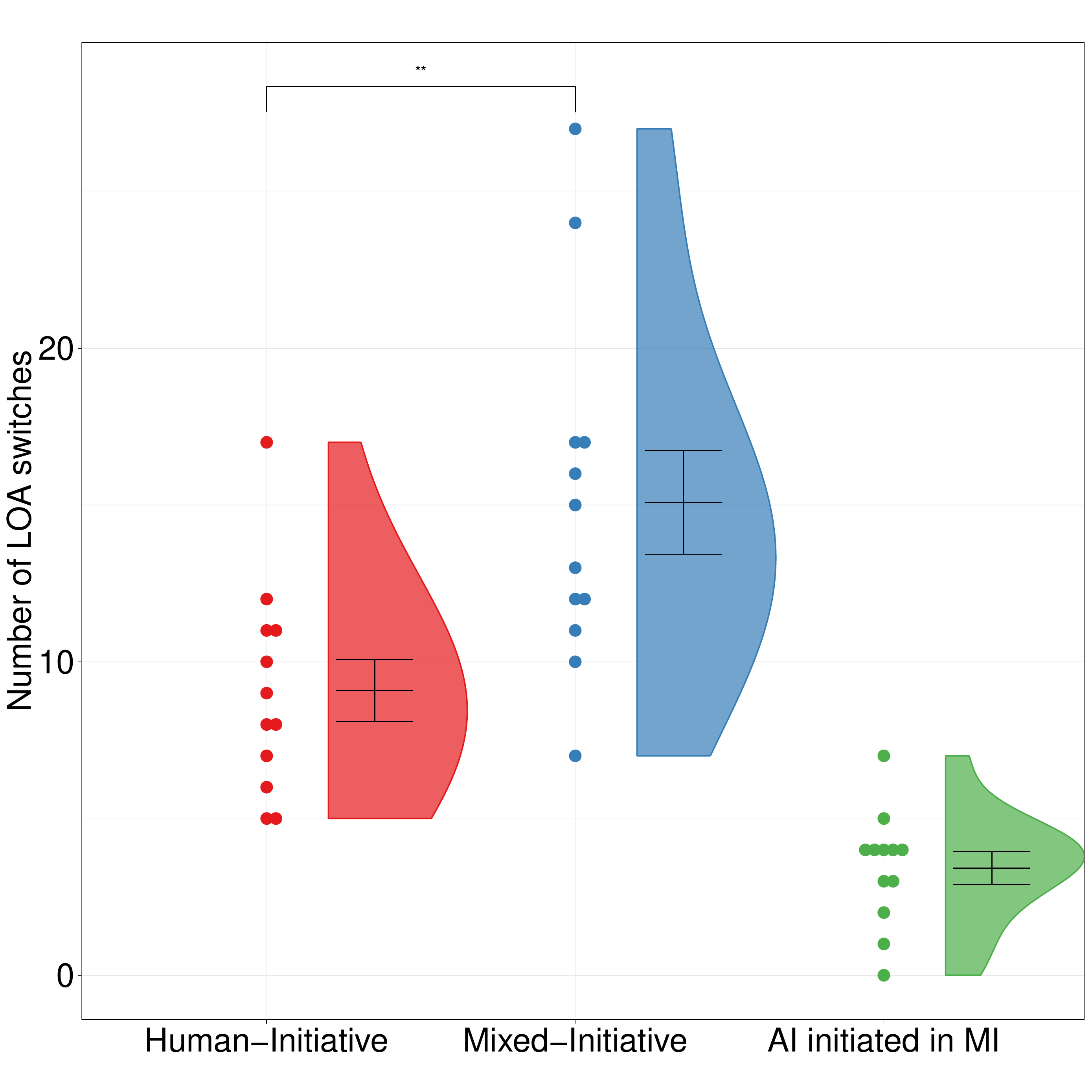}
			\caption{}
			\label{subfig:loa_exp3}
		\end{subfigure}
		\hfill
		\begin{subfigure}[b]{0.49\textwidth}
			\centering
			\includegraphics[width=\textwidth]{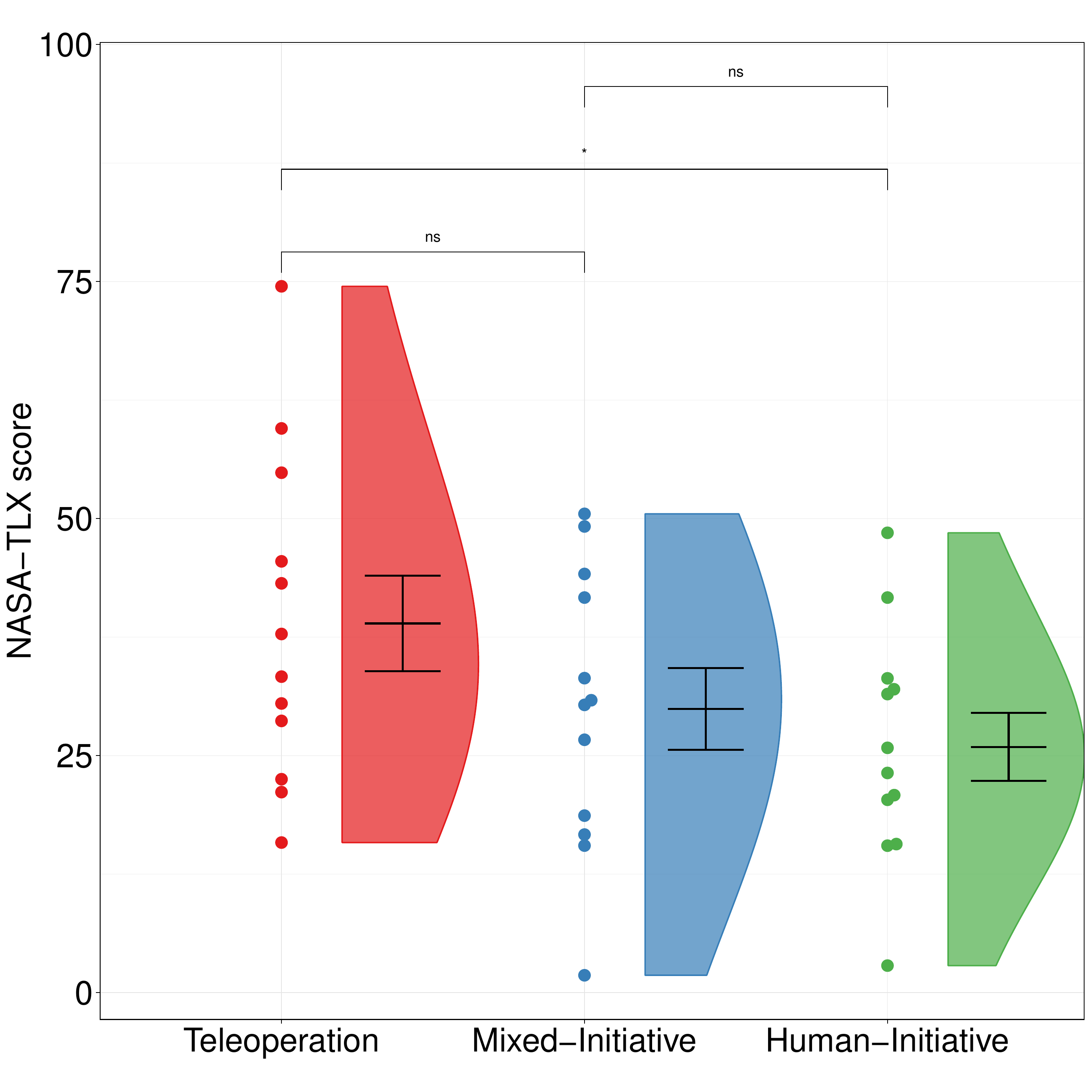}
			\caption{}
			\label{subfig:nasa-tlx_exp3}
		\end{subfigure}
		\hfill
		\caption{\textbf{\ref{subfig:loa_exp3}:} The number of LOA switches during HI, MI, and EMICS initiated LOA switches during MI. \textbf{\ref{subfig:nasa-tlx_exp3}:} NASA-TLX score showing the overall trial difficulty-workload as perceived by the operators.}
		\label{fig:exp3_tlx_loa}
	\end{figure*}

Control mode had a significant effect on \textit{NASA-TLX scores} (see Figure \ref{subfig:nasa-tlx_exp3}) as suggested by ANOVA. Pairwise comparisons showed that teleoperation was perceived as harder (i.e. more workload) compared to HI with $p < 0.05$ and marginally harder than MI with $p = 0.05$. HI variable autonomy is perceived as having the same difficulty as MI ($p > 0.05$). 

The mean \textit{number of LOA switches} in HI and in MI were compared using a paired samples t-test (see Figure \ref{subfig:loa_exp3}). The reason for using a paired samples t-test is that the experiment had a within-groups design, i.e. we expect some correlation in the results given that the same participants performed both conditions. The mean number of LOA switches in MI was significantly higher than the number of LOA switches in HI, as shown by the t-test. Using Pearson's correlation, no correlation was found in mean LOA switches between HI and MI ($r(10) = .472, p >. 05$). Given that the correlation assumption of paired samples t-test does not hold true, we used a independent samples t-test to validate the result further. Again, the means of HI and MI are significantly different. No correlation was found between the \textit{primary task completion time}, \textit{secondary task completion time}, \textit{NASA-TLX score} and the \textit{number of LOA switches} in MI.  

The mean number of LOA switches in MI due to EMICS's initiative is \textit{$M = 3.42$, ($SD = 1.83$)}. This number of LOA switches due to EMICS's initiative is positively correlated ($r(10)= .681, p<.05 $) with the \textit{primary task completion time}. No correlation was found between the \textit{secondary task completion time}, \textit{NASA-TLX score} and the mean number of LOA switches in MI due to EMICS's initiative.


\subsection{Discussion}
Regarding performance in the primary task, the HI control switcher outperformed teleoperation both in terms of score and time-to-completion. This evidence further reinforces our prior findings \cite{Chiou2016} that HI variable autonomy outperforms individual LOAs such as teleoperation and autonomy. Of particular importance is that this evidence came from conducting a scenario in which we aimed for realism and did not restrict potential degradation factors (e.g. naturally occurring noise or communication issues), while at the same time having a controlled experiment. 

In contrast to our previous experiment (see Section \ref{section:experiment2_2}), results from the use of MI control switcher are more difficult to interpret. A trend can be seen as MI performed better than teleoperation in the primary task, however the result was not statistically significant. A comparison between HI and MI does not lead to reliable conclusions, as no statistical difference was found. We believe this is due to conflicts between the operator's and the EMICS's initiative regarding LOA switching. This conflict arose from a restriction that the MI control switcher has. It is designed on the assumption that the agent who is in control (e.g. the MI system or the operator) follows the path yielded from the expert planner relatively closely. Our experiment was designed to control for exploration strategies by restricting operators in only visiting the center point of the offices/areas, as victims and hazards were visible from that point. However, there were occasions that operators decided to engage in some exploration or follow a less restricted path (i.e. compared to the one the expert planner yielded). For example operators that decided to move closer to a hazard sign in order to see the letters more clearly or in order to improve lighting conditions on the sign. Another example is provided by the operators that struggled to pass through the narrow passage created by the unseen object. In such cases the EMICS inferred a performance drop or a deviation from the navigation goal; as a result the EMICS switched to autonomy. At the same time, if operators had not yet finished their action, they would switch back to having control (i.e. teleoperation). This is further reinforced by participants' feedback. Many of them noted that they generally trusted the EMICS as there were cases in which it switched LOA in a meaningful way. However, they felt restricted from driving freely and they also felt the EMICS was intrusive at times. Others noted that they felt the MI capabilities were redundant, given that HI control switcher was easy enough to use.

Every time a LOA switch takes place, the EMICS re-initializes the error exponential moving average for a time period of a couple of seconds. During this period the control switcher cannot initiate a LOA switch. In essence, this period of time acts as a minimum time between EMICS initiated switches. Despite this period, the conflict for control was not avoided, suggesting that it is happening on a longer time scale (e.g. several seconds). Instead, this conflict for control between the EMICS and the operator can be avoided to a large extent by MI control switchers that are context-aware. Imagine the following two situations and a MI control switcher that is not context-aware. In the first situation the robot is idle (i.e. no progress towards the goal) because the operator is neglecting it in order to perform a secondary task. The performance error measured by the control switcher is the maximum. In the second case the robot is stuck in a corner and the operator is reversing in order to escape the enclosure. Similar to the previous situation the performance error is the maximum. An MI control switcher that does not take context into account, would initiate a LOA switch in both cases. In the first case the switch into autonomy would be beneficial as the robot was idle. In the second case it can potentially lead to a collision as the operator is in the middle of a maneuver. It can also lead to a control conflict as the operator can try to take control back in order to complete the maneuver. A context-aware control switcher would have been able to distinguish that although in both cases the performance error was large, the situation was different. Hence, in the second case a LOA switch wouldn't have happened. Our EMICS in this specific example, is aware that the large error while the robot is reversing potentially means maneuvering to escape an entrapment (see fuzzy rule base in Section \ref{fuzzy_controller}). However, with cases such the ones discussed in the previous paragraph (e.g. operator performing exploration), our control switcher is not able to cope. We identify context awareness as a major challenge for robotic MI systems. We believe that fuzzy control switchers can help towards this as they can be expanded with expert knowledge regarding context. Lastly, system transparency is another factor that might positively contribute towards tackling LOA switching conflicts. 

The above evidence suggesting a control conflict is in accordance with the fact that MI had significantly more LOA switches compared to HI. In contrast to the results of Section \ref{section:experiment2_2} in which LOA switches of MI and HI were highly correlated, in the experiment reported here no correlation was found. This further reinforces the notion that possibly these extra LOA switches stem from the robot-operator conflict for control as they would switch LOA back and forth. Similarly, a number of LOA switches initiated from the EMICS might be due to this control conflict. Possible evidence might come from the positive correlation found between the mean number of EMICS's initiated LOA switches in MI and the primary task completion time. The more LOA switches the EMICS initiated, the more time it took participants to complete the primary task. This is either an indication of a conflict for control degrading performance, or it means that the EMICS is more active in switching LOA for the under-performing participants (i.e. they need more help from the EMICS).

Further HRI studies are required in order to investigate the phenomenon better as this conflict is a major challenge to overcome for MI systems. Lastly, the fact that the number of LOA switches is relatively high for both HI and MI, is in accordance with the findings of our previous two experiments (see \cite{Chiou2016_AAAI} and Section \ref{section:experiment2_2}). These findings suggested that the high number of LOA switches was due to reasons beyond performance (e.g. personality traits).

The secondary task performance (i.e. time-to-completion and number of errors) were on the same level for all three control modes. This evidence suggests that LOA switching capabilities did not have any effect on the secondary task performance. This is similar to the evidence of our previous work \cite{Chiou2016}, in which teleoperation and HI performed equally on the secondary task. It also reinforces the possibility that the improved secondary task performance on our initial MI control switcher evaluation (compared to HI - see Section \ref{section:experiment2_2}), was due to learning effects. However because: a) the secondary task on the previous experiment is different from our current task; and b) the statistical power (i.e. the probability that a significant difference will be found if it exists) on our number of errors calculations is low; a reliable conclusion cannot be reached.

Regarding the difficulty/workload of the trials, NASA-TLX showed that teleoperation was perceived as the most difficult control mode compared to HI and MI. This suggests, similar to the findings of \cite{Chiou2016}, that the use of variable autonomy can alleviate operators from the burden of control. Perceived difficulty for MI and HI was on the same level according to the pairwise comparisons. This is in contrast to the findings of Section \ref{section:experiment2_2} in which MI found to be easier than HI. However, due to the low statistical power, the possibility of MI and HI differing is not excluded.

Lastly, a possible explanation for performance and workload not been fully in agreement with the first experiment can be given by the model of Johnson et al. \cite{Johnson2017} (similar to the discussion in Section \ref{section:learning_discussion}). At times, operators were allocating much of their cognitive resources to deal with the conflict for control. Hence, part of the performance degradation and workload arises from not been able to re-allocate their attention efficiently to the tasks.

\section{Limitations, insights and future work}

The main limitation of the expert-guided MI control switcher presented here is that is not able to cope with the conflict for control. Two of the assumptions on which the EMICS was based are: a) the human operator is willing to be handed or to hand over control based on the EMICS initiative; and b) the agent to which the control will be handed is capable of correcting the task effectiveness degradation. A conflict for control arises when one or both of these assumptions are violated. The main reason for those assumptions being violated was the miss-match at times between what the two agents considered ideal performance. For example, the human was acting with the aim of improving exploration performance or avoiding an unforeseen obstacle, while the MI control system was trying to optimize for a different trajectory between known waypoints and without any context knowledge. Given the evidence presented in detail in the previous sections, we consider this conflict for control the main factor affecting the performance of MI in the second experiment compared to that of the first simulated experiment.

In our first simulated experiment both the primary and the secondary tasks were more restricted compared to the tasks in the second real world experiment. Hence, the operators' actions remained within the experiment's EMICS's working envelope (i.e. relatively simple navigation). This resulted in the EMICS being able to generate a timely LOA switch when needed; and most importantly it did not try to switch LOA when that would be unnecessary or intrusive. This in turn allowed the operators to trust the EMICS. Hence, the two above assumptions held true. In our second physical robot experiment operators in many cases deviated from the navigational goal which they had given to the MI control system in order to explore further or move freely. The EMICS lacked the knowledge of the new goal (e.g. next to the hazard sign which the operator wanted to inspect instead of the center of the office), or the context of operator's commands. Due to this lack of knowledge the EMICS would initiate a LOA switch believing that the operator was not moving towards the goal, while instead the navigation goal had been changed. The operators found this intrusive and hence switched LOA, leading to the violation of both assumptions and to performance degradation.  

In essence the conflict for control arises because the EMICS and the operator have the authority to aggressively override each other's actions even in cases when they are not necessarily correct. When the EMICS incorrectly overrides the operator's actions it is due to: a) an incorrect inference about the operator's intention (e.g. lack of knowledge of the current goal); or b) a lack of contextual knowledge regarding the task and the operator's action (i.e. in which context a specific action is taking place and what this action is trying to optimize). When the operator overrides EMICS actions in the problematic cases, is mostly because they perceive the EMICS initiative as intrusive and aggressive and hence they do not want to hand over control. This is similar to the findings of Dragan and Srinivasa \cite{Dragan2013} in shared control. They found that when the AI controller is aggressive (i.e. intrusive) and wrong (i.e. takes a wrong action) it negatively affects performance and also user preference. However, in shared control any potential conflict for control is of a different nature as the operator cannot directly switch LOA and take over.

Regarding the expert-guided MI control approach we propose in this paper, further investigation is needed to tackle some of the confounding factors. More specifically a new experiment could be conducted in a real environment focused on comparing HI with RI and MI without a teleoperation condition. The experimental design should: a) use a higher number of participants; b) use a bigger robot arena; c) adjust the primary and secondary tasks to make them more difficult. Particular attention should be given to how to minimize the conflict for control either via the experimental design or via extending the MI control switcher.

An important direction for future work is how to make MI control switchers more aware of context. The evidence presented in this paper suggests that this would positively impact the conflict for control. A way to provide context is a MI control system capable of predicting operator's intent. For example, predicting operator's intent in the context of navigation (i.e. the navigational goal which the operator wants to move to) can inform the MI control switcher for a change in task context, e.g. a new navigational goal can mean that the operator started exploration. In essence, context in the form of operator's intent, can provide the MI control switcher with the ability to understand that its expert policy for ideal performance does not apply in a specific situation. Intent prediction is also known to improve performance in shared control manipulation tasks \cite{Jain2019}. Another way to provide context is a MI control system capable of inferring or taking into account the operator's status. For example, understanding if performance degradation is due to a tired operator \cite{Chanel2020} or if the operator is simply not available due to a secondary task \cite{Gateau2016} can improve the MI control switcher's assumptions about the capability and availability of the agent to which the control will be given.

Another direction for future research is improving the hand-off strategy of the MI control switcher (i.e. the way control is transferred between agents). Using a negotiation interface, e.g. similar to negotiation theory used to negotiate evasive maneuvers in autonomous car research \cite{Rothfuss2019}, can potentially enable smooth LOA switches as both agents can negotiate their choice contrary to the current aggressive override. For example an operator might refuse to hand over control in a middle of a maneuver and similarly the MI control switcher might insist on handing over control in a critical situation. This will improve the assumption that the agent is always willing to hand over or be handed control. Additionally, transparency about the MI control switcher's choice (i.e. communicating to the operator the reasons behind a LOA switch) can further facilitate smoother hand-offs.

Lastly, a mathematical formalism for MI control in the context of LOA switching, similar to the one of Dragan and Srinivasa for shared control \cite{Dragan2013} is needed. This, in conjunction with the MI control taxonomy of \cite{Jiang2015}, would give a solid foundation and a common understanding for future research.

\section{Conclusion and impact}

This paper presented an expert-guided approach to designing robotic MI control systems. It also presented an expert-guided Mixed-Initiative control switcher and its experimental evaluation using both a high fidelity simulator and in a realistic scenario.

The proposed control switcher uses an online performance metric that represents the expected effectiveness of goal directed motion, with parameters determined by comparing the control switcher to human performance on a prior experiment. The proposed control switcher aims to measure human-robot team performance, infer if a LOA switch is needed, and switch LOA. 

Evidence from our initial evaluation showed some of the potential advantages of MI control. The EMICS was found in simulation to outperform HI control in both primary and secondary task performance and in the perceived by the operator workload. This in turn means that MI outperforms control modes that lack any LOA switching capabilities such as teleoperation and autonomy.

The second evaluation extended our experimental framework towards a less controlled and more realistic setting. The EMICS was used in a real robot performing in a USAR scenario. Results regarding performance advantages of MI over HI were less conclusive compared to the initial evaluation. However, the experiment yielded significant new insight into challenges and problems which need to be overcome in the design of MI systems. The difficulty in interpreting the results was partially due to variance and poor statistical power in some of the statistical calculations. However, the major factor that affected the results was the conflict for control situations that arose between the operator and the EMICS. This led operators and the EMICS to confusion on what they should do as the LOA was switching back and forth between autonomy (i.e AI autonomous navigation in control) and teleoperation (i.e. human in control). We believe this control conflict is one of the major challenges for future MI control research and that the addition of further context-awareness (of task, environment, workload, operator's intent etc.) is one way to address this problem. However, more generally, MI control has its merits since it provides redundancy in the cases where an operator might not be able to switch LOA if needed e.g. during loss of communication with the robot, or a sudden event impairing the operator. 

Lastly, the USAR experiment provided important real world evidence that variable autonomy control in the form of HI has the potential to outperform teleoperation in a navigation task. This is in accordance to our previous findings in \cite{Chiou2016} and denotes that human operators successfully use HI capabilities to overcome various performance-degrading factors and situations. Additionally, further evidence was provided to support the hypothesis that operators switch LOA for reasons other than performance. 

Overall, we believe that this paper has made a number of significant contributions to the MI research on mobile robots: a) proposed a framework for designing MI control systems; b) proposed an MI control switcher (i.e. the EMICS); c) highlighted fields of interest and situations that would benefit from MI control while identifying the shortcomings that constitute open challenges for the research field.

\section*{Acknowledgment}
This work was funded by the British Ministry of Defence via the Defence Science and Technology Laboratory (Dstl), under their PhD bursary scheme, contract no. DSTLX-1000074621. It was also supported by the UK's Engineering and Physical Sciences Research Council (EPSRC) under grants: the National Centre for Nuclear Robotics, EP/R02572X/1; and related grants EP/M026477/1, EP/P017487/1 EP/P01366X/1. Rustam Stolkin was partly funded by a Royal Society Industry Fellowship.

\bibliography{IEEEabrv,refs}
\bibliographystyle{IEEEtran}

\end{document}